\documentclass{ieeeaccess}
\usepackage{booktabs}
\usepackage{float}
\usepackage{booktabs}
\usepackage{cite}
\usepackage{amsmath,amssymb,amsfonts}
\usepackage{algorithmic}
\usepackage{graphicx}
\usepackage{textcomp}
\usepackage{algorithm}  
\usepackage{gensymb}
\usepackage{lipsum,amsmath}
\usepackage{cuted}
\usepackage{bm}
\usepackage{multirow}
\usepackage{subfigure}
\usepackage{gensymb}
\usepackage{multirow}
\usepackage{tabularx}
\usepackage{array}

\usepackage{CJK}

\begin{document}
\history{Date of publication xxxx 00, 0000, date of current version xxxx 00, 0000.}
\doi{}

\title{Communication Efficiency Optimization of Federated Learning for Computing and Network Convergence of 6G Networks}
\author{\uppercase{Yizhuo Cai}\authorrefmark{1}, 
\uppercase{Bo Lei}\authorrefmark{2}, 
\uppercase{Qianying Zhao}\authorrefmark{2},
\uppercase{Jing Peng}\authorrefmark{3},
\uppercase{Min Wei}\authorrefmark{2},
\uppercase{yushun zhang}\authorrefmark{1}, and XING ZHANG\authorrefmark{1}}

\address[1]{Wireless Signal Processing and Network Laboratory,
Beijing University of Posts and Telecommunications, Beijing 100876,
China}
\address[2]{Research Institute of China Telecom Co., Ltd., Beijing 102209, China}
\address[3]{
Beijing Telecom, No. 21 Chaoyangmen North Street, Dongcheng District, Beijing}

\tfootnote{ This work was supported by the National Science Foundation of China under Grant 62271062,62071063.}

\corresp{Corresponding author: Xing Zhang and Bo Lei (email: zhangx@ieee.org).}

\begin{abstract}
Federated learning effectively addresses issues such as data privacy by collaborating across participating devices to train global models. However, factors such as network topology and device computing power can affect its training or communication process in complex network environments.
A new network architecture and paradigm with computing-measurable, perceptible, distributable, dispatchable, and manageable capabilities, computing and network convergence (CNC) of 6G networks can effectively support federated learning training and improve its communication efficiency. By guiding the participating devices' training in federated learning based on business requirements, resource load, network conditions, and arithmetic power of devices, CNC can reach this goal. 
In this paper, to improve the communication efficiency of federated learning in complex networks, we study the communication efficiency optimization of federated learning for computing and network convergence of 6G networks, methods that gives decisions on its training process for different network conditions and arithmetic power of participating devices in federated learning. The experiments address two architectures that exist for devices in federated learning and arrange devices to participate in training based on arithmetic power while achieving optimization of communication efficiency in the process of transferring model parameters. 
The results show that the method we proposed can (1) cope well with complex network situations (2) effectively balance the delay distribution of participating devices for local training (3) improve the communication efficiency during the transfer of model parameters (4) improve the resource utilization in the network.
\end{abstract}

\titlepgskip=-15pt
\maketitle

\section{Introduction}
\subsection{context}
In federated learning, machine learning models are trained collaboratively on multiple devices without the need for third parties to share training data. Therefore, federated learning with distributed architecture can effectively solve the problem of data silos. Moreover, compared with traditional machine learning it reduces data transfer costs while ensuring fairness among participants. But there are still many challenges to federated learning. The main challenges are manifested in several major aspects such as data privacy, communication consumption, device
training latency, and statistical heterogeneity during the training process \cite{b1,b2,b3}. Among these, communication issue is a huge challenge.

But today, computing and network convergence (CNC) of 6G networks has been proposed as a new network architecture and paradigm. Its various characteristics can fit well with federated learning. In detail, the CNC can rely on powerful heterogeneous resource scheduling capability and ubiquitous network connection to achieve real-time, flexible,  and efficient resource allocation, while ensuring synchronous information sharing of the whole network.  
Its efficient information-sharing mechanism, high utilization of computing power resources of nodes in the network, and the ability to select and schedule network data traffic paths can effectively improve the data and information communication efficiency of participating clients in the federated learning and optimize the whole training process. This shows that the combination of federated learning and CNC is necessary and effective.
\subsection{related works}

Solutions to the communication problem in federated learning focus on several aspects, such as reducing the amount of data for transmitting model parameters, reducing the frequency of communication, changing the type of communication, changing the aggregation method, using routing strategies, employing knowledge distillation, etc. As the model parameters or gradients passed in federated learning exist in the form of matrices, some scholars have used compression, quantization, and sparsification to reduce the amount of data while ensuring the validity of the data. Based on this idea they adopted two methods of random number seeding and sketch updating to reduce the amount of data for the model parameters\cite{b4}.

Meanwhile, McMahan \emph{et al.} \cite{b5} reduce the communication frequency by performing a global aggregation only after the participating clients have performed several local training times. Moreover, Using sampling to select some clients is able to reduce the communication consumption required to transfer the model parameters in each global training \cite{b5,b6}. However, Fraboni \emph{et al.} \cite{b6} divided the clients into different categories according to their local data distribution. Then, They sample clients for each global training from different categories, which is better than the former method.

There is also a problem of dropouts in communications challenges. Wu \emph{et al.} \cite{b7}  proposed to change the form of communication into semi-asynchronous communication. Moreover, there is one way to select participating clients for each round using a proportional fair scheduling strategy to reduce the possibility of dropouts \cite{b8}.

Another effective way to address communication is to change the approach of global aggregation \cite{b9,b10,b11}. So \emph{et al.} \cite{b9} focus on secure aggregation, while others introduce aggregation servers at the edge to reduce communication consumption \cite{b10, b11}. And as the popularity of knowledge distillation grows, knowledge distillation is also used for federated learning. 
Based on the idea of knowledge distillation, only the predicted logits values or the parameters of the student model are transmitted in the federated learning \cite{b12,b13,b14}. This significantly reduces the amount of data to be transferred.

Dinh \emph{et al.} \cite{b15} derived the convergence analysis and resource allocation problem of federated learning in wireless networks.
On the other hand, Chen \emph{et al.} \cite{b16}  accelerate model convergence by reducing data transmission errors of federated learning clients in a wireless environment. And Qin \emph{et al.} \cite{b17} also proposed specific scenarios for federated learning in wireless environments.
Moreover, using a multi-layer aggregation approach to reduce the consumption of bandwidth resources in a specific tree structure can reduce the latency of federated learning model aggregation \cite{b18}.

In addition, novel machine learning-enabled wireless multi-hop federated learning framework can greatly mitigate the adverse impact of wireless communications on the federated learning performance metrics \cite{b19}.

However, the heterogeneity of device computing power in federated learning can seriously affect its communication process. As different computing power leads to different local training latency, it quite affects the subsequent communication process. Typically, traditional third-party aggregation servers tend to be insensitive to the perception of information such as the computing power of the device. Sun \emph{et al.} \cite{b20} investigated a future network paradigm capable of connecting distributed computing nodes. It's able to dynamically and timely sense user needs and multidimensional resources such as applications, network resources, computing power resources, and storage resources. The increased efficiency of federated learning communications can take advantage of exactly these properties.

\subsection{contributions}
Inspired by the reference \cite{b20}, we propose communication efficiency optimization of federated learning for Computing and Network Convergence of 6G Networks. The CNC of 6G Networks is a new network architecture and paradigm with greater computing-measurable, perceptible, distributable, dispatchable, and manageable capabilities. Relying on the ability to sense the computing power of client devices in real-time and synchronize various resource information in the network, the CNC can flexibly schedule the clients in federated learning and make better decisions for the network topology and the allocation of the resources. This guides the training process of federated learning and improves communication efficiency.

The main contributions of this paper are summarized as follows:
\begin{enumerate}

\item We propose a federated learning system of the CNC and its communication efficiency optimization methods. Combined with the research known, we divided the CNC into different layers. When each layer interacts with each other, CNC senses the network condition and client devices resources, and makes decisions to optimize the federated learning training process.

\item As two architectures exist for federated learning training, we have adopted corresponding optimization methods. All of them improve the performance of federated learning.

\item We focus on the computing power heterogeneity of client devices. By scheduling the clients in each global training the computing power difference between clients is balanced. Under the traditional architecture, the average delay difference of the proposed system per training round is one-fifth of the federated averaging (FedAvg) algorithm in \cite{b5}. The maximum value is about 46.6\% of FedAvg. Under the peer-to-peer architecture, by scheduling the computing power of the devices, the model accuracy of our system converges faster with the same training time than the baseline.

\item Federated learning communication efficiency optimization for CNC of 6G networks could improve the resource utilization of the whole network.
During federated learning, optimization methods can update the information and schedule the resources about client devices or the network in real time. With this support, in the traditional architecture of federated learning, the transmission latency and energy consumption per round of model training are reduced by about 46.9\% and 19.4\% compared to FedAvg\cite{b5}. In peer-to-peer architecture, the performance of the system regarding transmission problems varies with the network environment, but is similar to, and sometimes better than the performance of the baseline.
\end{enumerate}

The rest of this paper is organized as follows. Section \uppercase\expandafter{\romannumeral2}
introduces the system architecture of optimization for CNC of 6G networks. Section \uppercase\expandafter{\romannumeral3} explains the mathematical problems faced in optimization. Section \uppercase\expandafter{\romannumeral4}  proposes algorithms of optimization in two architectures. And section \uppercase\expandafter{\romannumeral5} 
shows the results of the experiment.
Section \uppercase\expandafter{\romannumeral6} summarizes our work.

\section{ Optimization system  }
\subsection{two architectures of federated learning}
\begin{figure}[htbp]
\centering
\subfigure[traditional architecture]{
\includegraphics[scale=0.35]{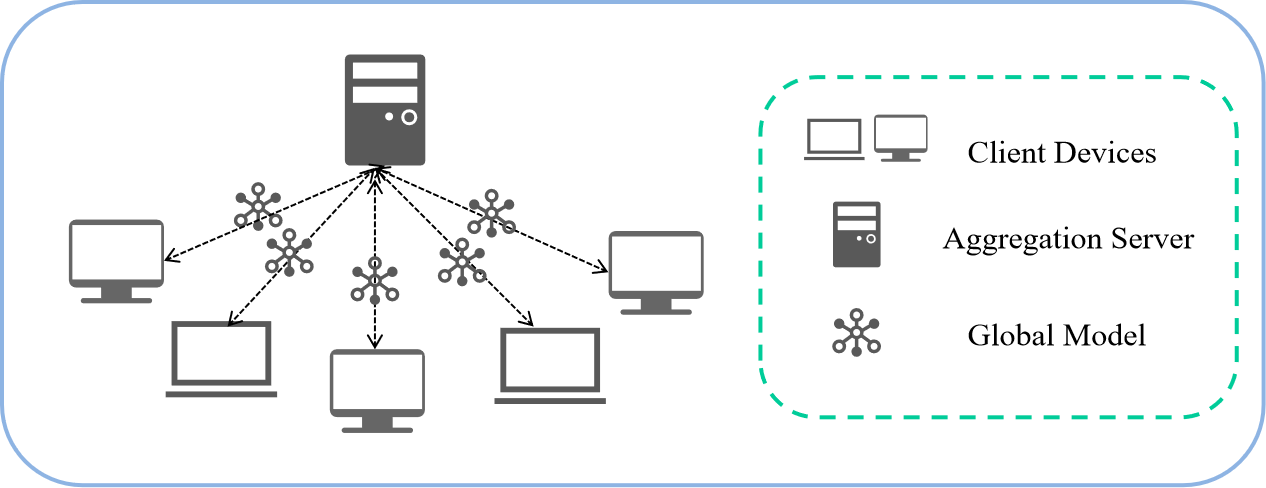} \label{fig1(a)}
}
\quad
\subfigure[peer-to-peer architecture]{
\includegraphics[scale=0.35]{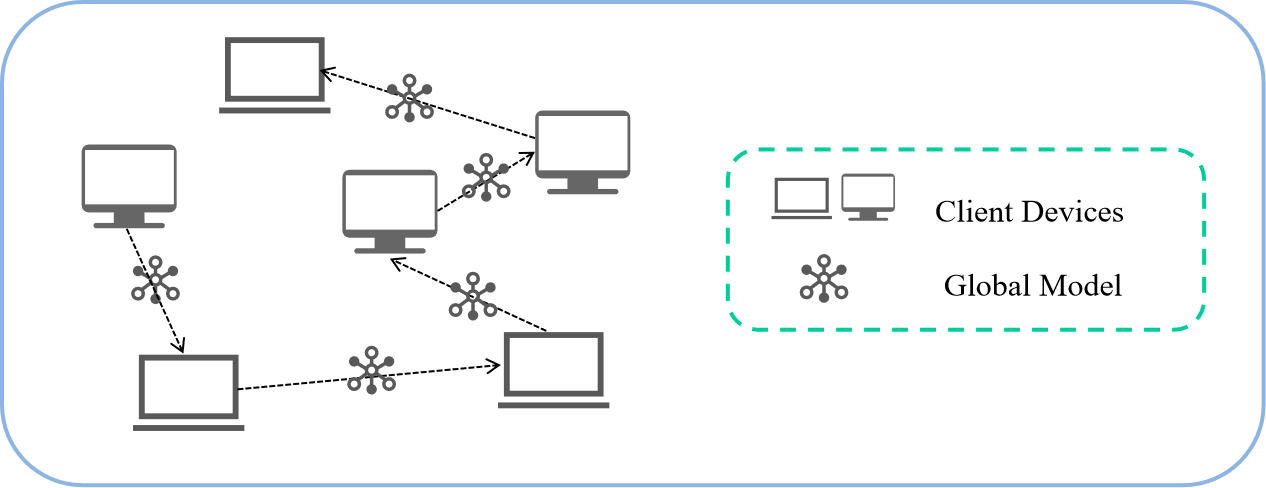} \label{fig1(b)} 
}

\caption{two architectures of federated learning}
\end{figure}

There are two training architectures in federated learning, as shown in Fig. \ref{fig1(a)}. In each global training, clients in traditional architecture receive the global model from the server and train it with their local data. Then they will transmit the updated model to the server.  After receiving the local models from all clients, the server performs an aggregation algorithm to obtain a new global model. Thus federated learning moves on to the next round of global training, which does not end until the global model reaches a certain accuracy.

In contrast, in a peer-to-peer architecture, federated learning exists only for the client that trains the model. All clients train and transmit the models in a specified order, forming a chain structure, as shown in Fig. \ref{fig1(b)}. The process in which the global model is transmitted from the initial client to the last client constitutes a global training round.

\subsection{Communication Efficiency Optimization system for CNC of 6G Networks}
\begin{figure*}[htbp]
    \includegraphics[width=\linewidth,scale=1]{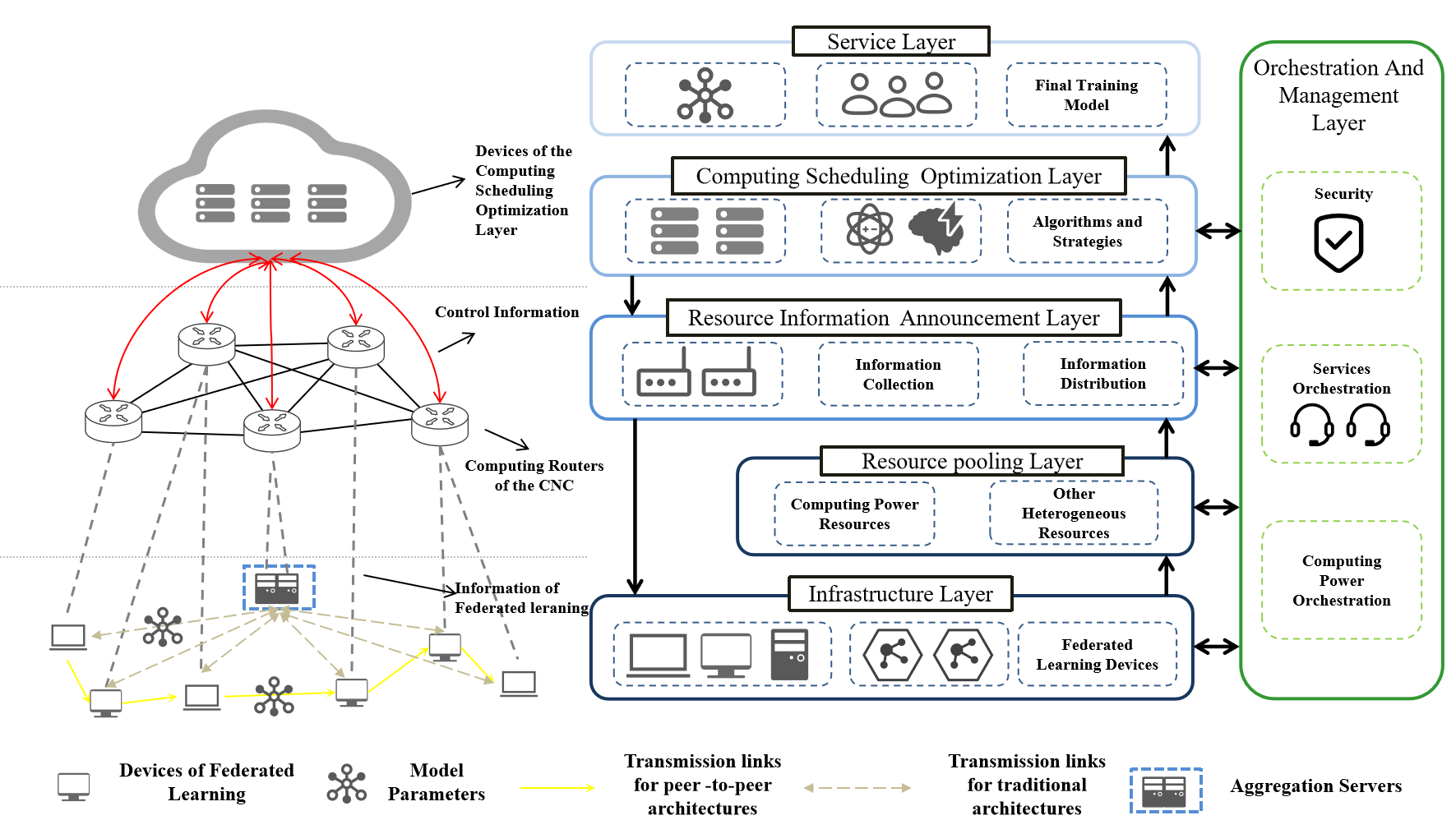}
    \caption{Communication Efficiency Optimization system for CNC of 6G Networks}
    \label{fig2}
\end{figure*}

Based on the two architectures, we propose communication efficiency optimization for CNC of 6G networks. The communication efficiency in federated learning training can be improved by exploiting computing power resource modeling, information synchronization, and scheduling capabilities of the CNC. Fig. \ref{fig2} shows its architecture.

\begin{figure*}[htbp]
    \includegraphics[width=\linewidth,scale=1]{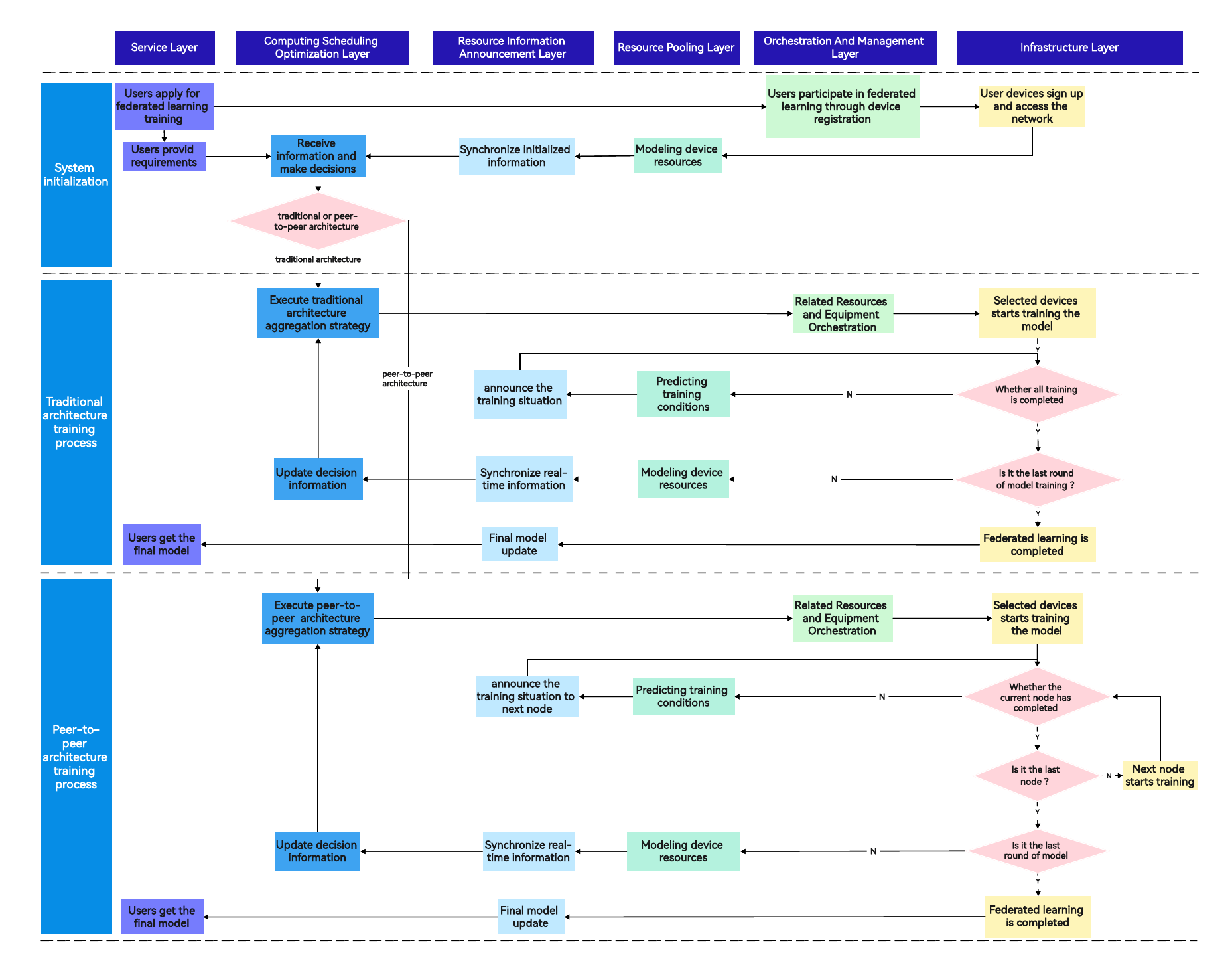}
    \caption{System operation flow chart}
    \label{fig3}
\end{figure*}

We stratify the CNC into an infrastructure layer, resource pooling layer, resource information announcement layer, computing scheduling optimization layer, service layer, and orchestration and management layer. As devices of the infrastructure layer, the aggregation servers and client devices involved in federated learning are scheduled and controlled by the CNC. Equipment in the resource pooling layer model the network resources, computing power resources, etc. of the underlying devices.
The information announcement layer, on the other hand, consists of specialized equipment. Downwards it collects various information from the participating devices or publishes training strategies. Upwards it forwards information about the clients to the scheduling optimization layer or obtains decision information from the upper layer. 
The scheduling and optimization layer is responsible for optimizing the federated learning scheduling algorithms and topological decisions based on the information from the underlying layer. Then it outputs the control information for the federated learning training in real time.
orchestration and management layer has control of the entire system of the CNC. It's responsible for orchestrating and scheduling the various resources used in federal learning, as well as managing the various devices in the other layers. The security and services orchestration layer is also under its control. 

The core of the whole structure is the coordinated operation of the upper equipment in the CNC, which optimizes the training process of federated learning based on the network, computing power resources, and other information. In Fig. \ref{fig2}, equipment of the scheduling optimization layer, these cloud servers, exchange the decision information of federated learning through the red line in the figure. And the routers of the information announcement layer are responsible for synchronizing the training information. They connect the whole network. Moreover, the infrastructure layer is divided into two cases. The traditional architecture requires the aggregation of server clusters and client devices to participate in the training, and the logical transmission topology is shown by the gray dashed line. In peer-to-peer architecture, only client devices are required to participate in training. The logical transmission topology of model parameters is given by the yellow line in Fig. \ref{fig2}.

Based on this structure, Fig. \ref{fig3} gives the overall flow of the whole system from initialization to entering the federated learning training process of both architectures and finally completing the training.

\section{Mathematics Problems}
The communication efficiency optimization of federated learning for computing and network convergence of 6G networks contains optimization work for several problems. This section introduces specific mathematical problems based on two architectures of federated learning.

\subsection{Convergence Problem}
Assume that there are $U$ participating clients. For client $i$, we use $D_i$ to denote the original data, matrix $\boldsymbol{X}_{\boldsymbol{i}}=\left[\boldsymbol{x}_{\boldsymbol{1}}, \boldsymbol{x}_{\boldsymbol{2}},  \ldots, \boldsymbol{x}_{\boldsymbol{n_i}}\right]$ to denote the feature vector matrix in $D_i$ that is the input of the model. $n_i$ is the amount of data of the client $i$. $\boldsymbol{Y}_{\boldsymbol{i}}=\left[y_{i 1}, y_{i 2},  \ldots, y_{i n_i}\right]$ indicate the corresponding labels of the client $i$ for the model output and $w_i$ is the local model parameters of each client. Federated learning is a process of continuous training models with data and iteratively aggregating them to find the best global model $w^*$. As the number of iterations increases, the global model gradually converges, approaching $w^*$. The goal of our optimization is to find $w$ so that it satisfies:
\begin{equation}
\min _{w_1, \ldots, w_i} \frac{1}{K} \sum_{i=1}^U \sum_{n=1}^{n_i} f\left(\boldsymbol{w}_{\boldsymbol{i}}, \boldsymbol{x}_{\boldsymbol{i n}}, y_{i n}\right).\label{eq1}\end{equation}

where $K=\sum_{i=1}^Un_i=\sum_{i=1}^U\left|D_i\right|, \text { and } w_1=w_2 \ldots=w_U=w$.
For the loss function$f()$, it can be defined according to the actual situation, such as the mean squared difference loss function.

\subsection{Transmission Problem}
The network topology and data transmission types of the clients under different architectures of federated learning determine the respective transmission problems.
\subsubsection{Transmission Problem under Traditional Architecture }
In this paper, we design a traditional architecture in which clients of the federated learning are in a wireless network environment. Each client occupies one Resource Block(RB) of the orthogonal frequency division multiple access technique. We follow the mathematical formulation of reference \cite{b16} to model the problem of client model parameter transmission. Based on this, the uplink rate at which the user transmits data to the central server can be expressed as:
\begin{equation}
r_i^U=B^U \mathbb{E}_{h_i}\left(\log _2\left(1+\frac{P_i h_i}{I_k+B^U N_0}\right)\right). \label{eq2}
\end{equation}

Where $B^U$ is the bandwidth of each RB.  $P_i$ is the transmitting power of client $i$. And $h_i=o_i d_i^{-2}$ is the channel gain between client $i$ and the central server. $d_i$ is the distance between the client $i$  and the central server.  $o_i$ denotes the Rayleigh fading parameter. Moreover, $\mathbb{E}_{h_i}()$is the expectation function for $h_i$. $I_k$ is the interference caused by other users using RB$k$ and $N_0$ is the noise power spectral density.

We assume that the consumption originates only from the data transmission process of the client device. So the transmission problem only needs to consider the uplink transmission delay and energy consumption of the clients:
\begin{equation}
l_i^{\mathrm{U}}=\mathrm{Z}\left(w_i\right) / r_i^{\mathrm{U}}.\label{eq3}
\end{equation}
\begin{equation}
e_i=P_i l_i^{\mathrm{U}}.\label{eq4}
\end{equation}

where $l_i^{\mathrm{U}}$ denotes the transmission delay, $e_i$ is the transmission energy consumption and $\mathrm{Z}\left(w_i\right)$ is the amount of data for the model parameters.

In each global training, clients are drawn from $U$, represented by the set $S_t$. Then the transmission problem can be interpreted as finding the best RB resource block allocation method that satisfies:
\begin{equation}
\min \sum_{i \in S_t} e_i.\label{eq5}
\end{equation}
or
\begin{equation}
\min \left(\max \left(\left\{l_i^{\mathrm{U}} \mid i \in S_t\right\}\right)\right).\label{eq6}
\end{equation}

\subsubsection{Transmission Problem under Peer-to-Peer Architecture }
In the peer-to-peer architecture, the transmission formula in the wireless environment is not applicable because there is no central server. We consider that all clients are connected to each other in the network. The transmission consumption and delay between the clients is denoted by ${cost}_{i, j}$. The transmission problem becomes: 
\begin{equation}
\min \left(\sum_{(i, j) \in  { trace\_path }} {cost}_{i, j}\right).\label{eq7}
\end{equation}

Where ${cost}_{i, j}$ is the transmission consumption between client $i$  and client $j$, and $trace\_path$ is the transmission order for selected clients.

\subsection{Local Training Related Problem}
For participating clients the local training delay is given according to the computing power of their devices. $c_i$ is the maximum computing power that can be output by the client $i$, $\left|D_i\right|$ is the amount of local data, and $epoch\_local$ represents the number of local training per global training. With the same model for each client, the local training delay can be roughly given as:
\begin{equation}
t_i=\alpha *  { epoch\_local } *\left|D_i\right| / c_i.
\label{eq8}
\end{equation}

Where $\alpha$ is the conversion factor of (\ref{eq8}). For federated learning, the local training problem can be solved by finding a suitable subset $S_t$ satisfying:
\begin{equation}
t_{\max }-t_{\min }<\varepsilon. \label{eq9}
\end{equation}

where $t_{\max }=\max \left\{t_i \mid i \in S_t\right\} $ ,
$t_{\min }=\min \left\{t_i \mid i \in S_t\right\} $, $\varepsilon$ indicates an acceptable time difference.

\section{Communication Efficiency Optimization of Federated Learning for CNC of 6G Networks}
According to the network topology and user requirements, we provide the optimization methods under two architectures of federated learning, as described below.
\subsection{Optimization under Traditional Architecture }

\begin{algorithm}[htbp]
\caption{Client scheduling strategy based on computing power}
\label{algoritem1}
\begin{algorithmic}[1]
  \REQUIRE{the set of participating clients $U$, the amount of data for peer client $\left|D_i\right|$,The computing power of participating device $c_i$, number of local training  ${ epoch\_local }$, conversion factor $\alpha$.}
  \ENSURE{the set of participating clients for each global training $S_t$.}
   \FOR{$i=1$ to $U$ }
   \STATE $ t_i=\alpha *  { epoch\_local } *\left|D_i\right| / c_i$
   \ENDFOR
\STATE Sort $\left\{t_i \mid i=1,2 \ldots, U\right\}$ by Descending Order 
\STATE Divide $U$ clients into $m$ parts, each part is denoted by $U_k$, $k=1,2\ldots,m$.
\STATE Sample from m parts,$P_k=\frac{N_k}{\sum_{k=1}^m N_k}$, where $N_k=\sum_{i \in U_k}\left|D_i\right|$.
\STATE Sample n clients from $U_k$ as $S_t$, $ P_i=\frac{\left|D_i\right|}{N_k}$
\RETURN $S_t$
\end{algorithmic}
\end{algorithm}
This optimization method utilizes all layers of the CNC to operate efficiently.
Based on FedAvg, we focus on the computing power problem and the model transfer problem. The system gives optimization based on these two points.

In federated learning, the clients register their local devices through the platform of the CNC to obtain a great global model. The devices of them are then scheduled by the devices in the orchestration and management layer for unified orchestration. Finally, the devices of clients operate as the infrastructure layer devices that support federated learning.

If the local data distribution of clients is similar, the choice of $S_t$ has little effect on (\ref{eq1}). However, its effect on (\ref{eq9}) is large. In our proposed algorithm, under the CNC of 6G Networks, the resource pooling layer devices will model the heterogeneous resources of client devices and upper-level equipment will group them based on the computing power. In this case, the clients selected for each global training have similar local training time and participating users can have a better experience. In each global training, The resource pooling layer facilities obtain the resource information of all devices and perform resource modeling. Devices of the resource information announcement layer forward these messages to devices in the computing scheduling optimization layer. Then the decision about $S_t$ will be made. The steps are given by Algorithm \ref{algoritem1}.

At the same time, we draw on the idea of reference\cite{b16}. By considering the consistent transmitting power of the clients, we make the optimization problem of communication performance an RB allocation problem. Devices in the computing scheduling optimization layer of the CNC execute Algorithm \ref{algoritem1} to obtain $S_t$ and then optimize the allocation strategy of resource blocks using the Hungarian algorithm. Different clients that are allocated different resources transfer data with different consumption. We construct the consumption matrix with the energy consumption generated by client $i$ transmitting data on RB $k$. The Hungarian algorithm often solves for a good solution about (\ref{eq5}).

Later, devices in the computing scheduling optimization layer hand the selected client information and the allocation policy of RBs to the resource information announcement layer devices for forwarding. And the clients in the federated learning can share the transmission performance, local training performance, etc. of other devices predicted by the CNC. 

The orchestration and management layer devices are based on algorithms to schedule and orchestrate federated learning devices and various resources. Then the participating clients receive the algorithm information output from the CNC to perform federated learning to obtain the converged global model by using the idea of weighted average. After getting the new model the system moves to the next iteration.

\subsection{Optimization under Peer-to-Peer Architecture }
Similarly, client devices operate as the infrastructure layer devices to support federated learning. The Algorithm \ref{algoritem2} shows the entire process steps.

Under Peer-to-Peer architecture, all clients are trained once as a global training. Before each global training devices in the resource pooling layer and resource information announcement layer cooperate to obtain information on computing power resources and network conditions of clients. Devices in the computing scheduling optimization layer assign subsets $S_{te}$ based on $c_i$ and $D_i$, where $e=1,2, \ldots, E$. For each $S_{te}$, the sum of local training delay is similar. Then they abstract the consumption matrix $G$  according to the network connectivity of each client. $cost_{i,j}$ represents the delay or energy consumption of the model transmission between the clients.

For clients in the $S_{te}$, the shortest transmission path over all clients needs to be found to solve (\ref{eq7}). With the powerful support of the CNC, we can generate the submatrix of $S_{te}$ from $G$ and then solve it by Algorithm \ref{algoritem3}.

\begin{algorithm}[t]
\caption{Optimization under peer-to-peer architecture}
\label{algoritem2}
\begin{algorithmic}[1]
  \REQUIRE{the set of participating clients $U$, the  data for peer client $D_i$.}
  \ENSURE{The final global model $w$.}

\STATE Participating clients and aggregation servers join The CNC of 6G Networks as node devices.
\FOR{$t=1$ to $T$}
\STATE devices in each layer of the CNC collaborate to divide the $E$ parts $S_{te}$. 
\STATE Calculate the transmission path for each subset $S_{te}$ according to Algorithm \ref{algoritem3}
\STATE Forward transmission strategy and initial model
\FOR{ client $i$ in $trace\_path$ of $S_{te}$ Parallel }
\IF{ client $i$ is the first client}
\STATE  Recieve model $w$ from the devices of the CNC , $w_i$ = $w$
\ELSE 
\STATE Recieve $w_{S_{t e}}$ from the previous client,$w_i$ = $w_{S_{t e}}$
\ENDIF
\FOR{$B$ in $D_i$}
       \FOR{$(x,y)$ in batch $B$}
          \STATE $w_i =w_i-\eta * \nabla f\left(w_i, x,y\right)$,
           $\eta$ is the learning rate.
       \ENDFOR
       \ENDFOR
\STATE $w_{S_{t e}}$ = $w_i$
\STATE Send the model $w_{S_{t e}}$ to the devices of the CNC if $i$ is the last client, otherwise next client
\ENDFOR
\STATE 
$w=\sum_{e=1}^E \frac{N_{t e}}{\sum_{e=1}^E N_{t e}} w_{S_{t e}}$, $N_{t e}=\sum_{i \in S_{t e}}\left|D_i\right|$
\ENDFOR
\RETURN $w$
\end{algorithmic}
\end{algorithm}

\begin{algorithm}[ht]
\caption{Optimal transmission path selection strategy}
\label{algoritem3}
\begin{algorithmic}[1]
  \REQUIRE{The consumption matrix of $S_{te}$:$G_e$. }
  \ENSURE{The optimal transmission path for$S_{te}$:$trace\_path$.}
   \FOR{$i$ in $S_{te}$ }
   \STATE Initialize  array variable $trace$, $trace=[[i]]$
\WHILE{$trace$}
 \STATE Get the last feasible path $current\_trace$, extract the last client $current\_point$ in $current\_trace$
\FOR{$j$ in $S_{te}$ }
\IF{client $j$ has been traversed or the client  is at infinite distance from the $current\_point$}
\STATE continue
\ENDIF
\STATE Save the next client $j$ and its distance to the client $current\_point$
\ENDFOR
\IF{no current saved next client}
    \STATE Remove the current path
    \STATE Continue
\ELSE 
\STATE Select the shortest distance of the client connected to $current\_point$ as the next client and output the latest transmission path
    \IF{all clients are traversed}
        \STATE Get $trace\_path$ of client $i$
        \STATE Break
    \ENDIF
    \STATE $trace$ stores the current feasible paths 
\ENDIF
    \ENDWHILE   
\ENDFOR
\STATE Select the path with the shortest sum of transmission consumption in $S_{te}$ as $trace\_path$
\RETURN $trace\_path$
\end{algorithmic}
\end{algorithm}

After outputting the transmission path for each subset $S_{te}$,  devices of the resource information announcement layer forward the transmission policy to all clients and passes the model to each initial client. The clients receive the information and start training and then transmit the parameters of the trained global model according to the policy. After each part is trained, the $E$ sub-models are aggregated by the specific devices in the computing scheduling optimization layer of the CNC to obtain the new global model and open the next round of training. Until the global model reaches a certain accuracy, other devices from the CNC, for instance, routers of the resource information announcement layer broadcasts the final global model to all clients.

\section{Simulation Results And Analysis}
In our experiments, we used the MNIST dataset, which can be independent and identically distributed (IID) or non-independent and identically distributed (Non-IID) after processing. We cut the datasets equally based on the total number of clients to get the clients with local data. Then we used a simple neural network as the training model for training.

\subsection{Simulation experiments under traditional architecture}
\subsubsection{Simulation environment parameter setting}

\begin{table}[htbp]
  \renewcommand{\arraystretch}{1}
  \caption{Parameter value setting under traditional architecture}
  \resizebox{\linewidth}{!}{
  \label{table1}
  \centering
  \begin{tabular}{|c|c|c|c|}
      \hline
      \bfseries Parameter & \bfseries Value & \bfseries Parameter & \bfseries Value \\
      \hline
      $N_0$ & -174 dBm / Hz & $m$ & 0.024 {~dB} \\
      \hline
      $B^U$ & 1 {MHz} &  $num\_clients$  & {[100,60]} \\
      \hline
      $P$ & 0.01 {~W} &  $cfraction$  & {[0.1,0.2]} \\
      \hline
      $I$ & {U}$(10^{-8}, 1.1 \times  10^{-8} )$ &   $local\_epoch$  & {[1,5]} \\
      \hline
      $d$ & {U}(0,500) & \ $batch\_size$  & 10 \\
      \hline
      $Z(g)$ & 0.606 {MB} & $lr$  & 0.01 \\
      \hline
      $o$ & 1 &  $global\_epoch$  & {[300,250]} \\
      \hline
  \end{tabular}
  }
\end{table}

Table \uppercase\expandafter{\romannumeral1} shows some parameter settings for the simulation experiments. In the experimental simulation under the traditional architecture, we use random number seeds to generate $I$ and $d$, and then calculate the communication transmission. Parameter $cfraction$ denotes the sampling proportion of each global training.

For the scheduling of computing power, the computing power of the local client device is equivalent to the local training time due to the consistent amount of client data as well as model size in the simulation. We tested a local training time of about 4s for a client. Then we set up the heterogeneous situation of the computing power resources of the clients and simulate the realistic situation after calculating the local training delay by (\ref{eq8}).

\begin{table}[htbp]
  \renewcommand{\arraystretch}{1}
  \caption{Abbreviation  for different cases}
  \resizebox{\linewidth}{!}{
  \label{table2}
  \centering
  \begin{tabular}{|c|c|}
      \hline
      \bfseries Abbreviation & \bfseries Parameter setting   \\
      \hline
      $Pr1$ & $num\_clients:100, cfraction:0.1, local\_epoch:1$   \\
      \hline
      $Pr2$ & $num\_clients:100, cfraction:0.1, local\_epoch:5$   \\
      \hline
      $Pr3$ & $num\_clients:100, cfraction:0.2, local\_epoch:1$   \\
      \hline
      $Pr4$ & $num\_clients:100, cfraction:0.2, local\_epoch:5$   \\
      \hline
      $Pr5$ & $num\_clients:60, cfraction:0.1, local\_epoch:1$   \\
      \hline
      $Pr6$ & $num\_clients:60, cfraction:0.1, local\_epoch:5$   \\
      \hline
  \end{tabular}
  }
\end{table}
In our experiments, we set up different cases to observe our approach. And we also compare performance with FedAvg \cite{b5}.
For convenience, Table \uppercase\expandafter{\romannumeral2} records the abbreviation corresponding to the different cases.

\subsubsection{experiment results and analysis}

\begin{figure}[htbp]
\centering
\subfigure[IID]{
\includegraphics[scale=0.5]{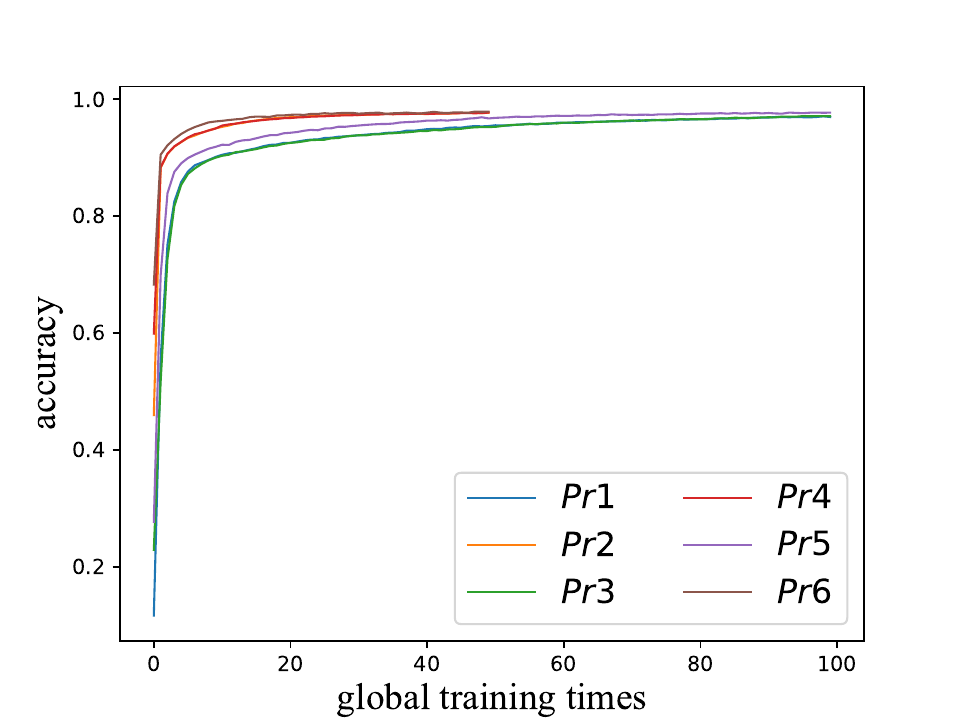} \label{fig4(a)}
}
\quad
\subfigure[Non-IID]{
\includegraphics[scale=0.5]{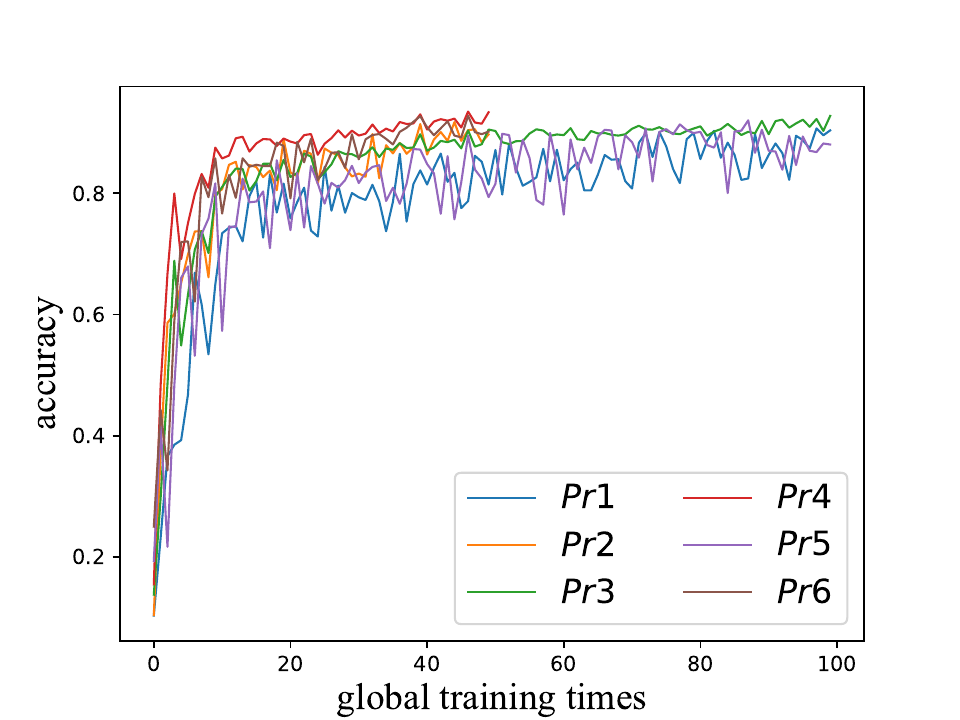} \label{fig4(b)} 
}

\caption{Results of global model test accuracy with communication efficiency optimization for CNC  under different cases}\label{fig4}
\end{figure}

\begin{figure*}[htbp]
\centering
\subfigure[transmission energy ]{
\includegraphics[scale=0.3]{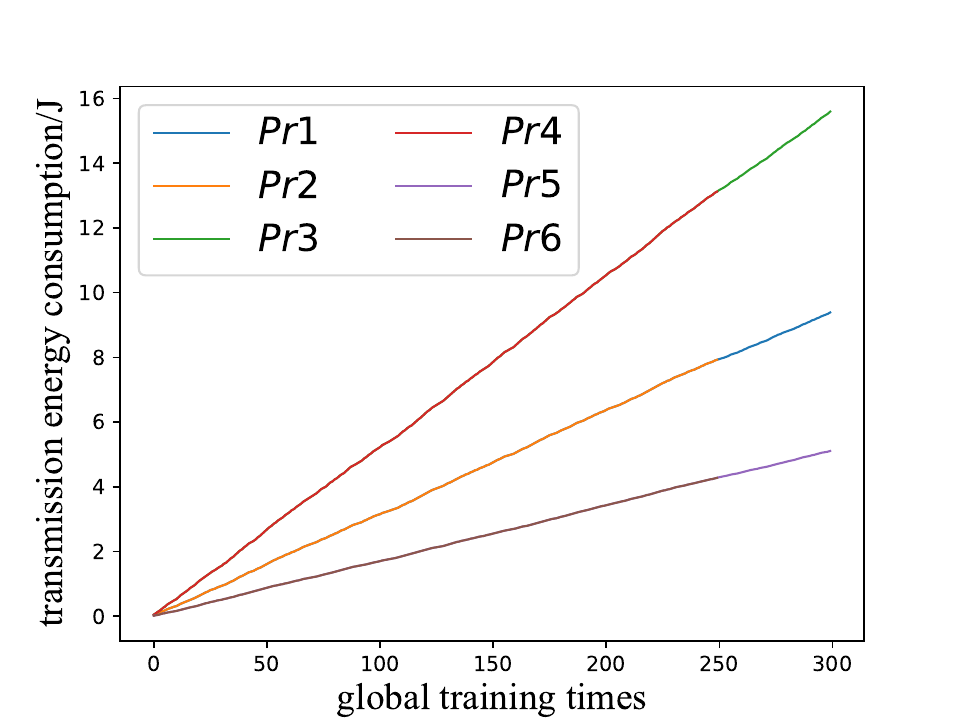} \label{fig5(a)}
}
\quad
\subfigure[transmission delay]{
\includegraphics[scale=0.3]{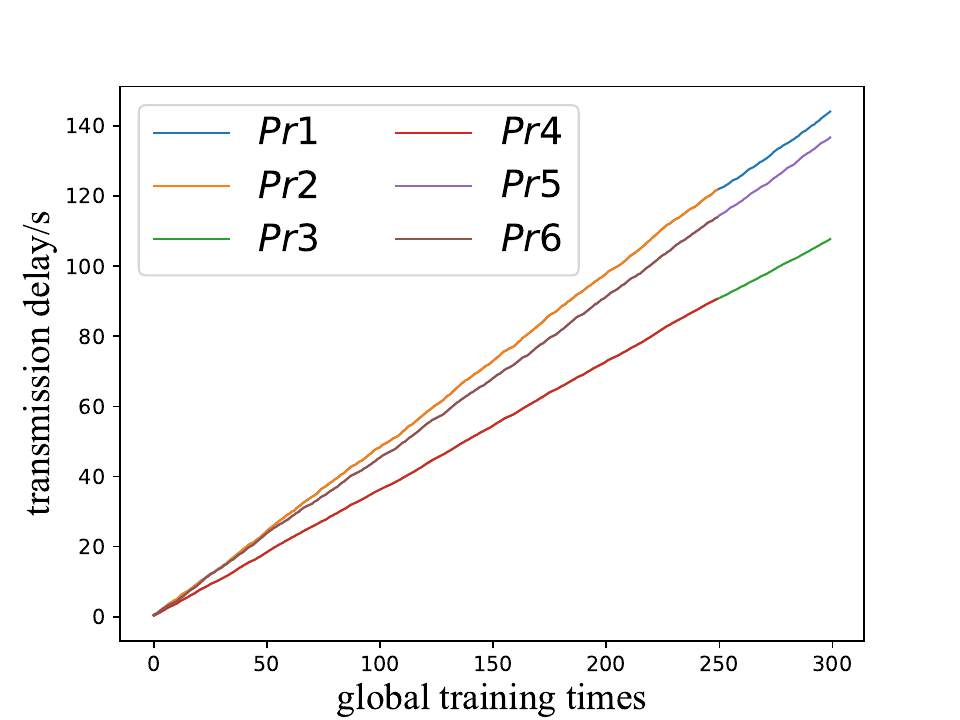} \label{fig5(b)} 
}
\quad
\subfigure[local training delay]{
\includegraphics[scale=0.3]{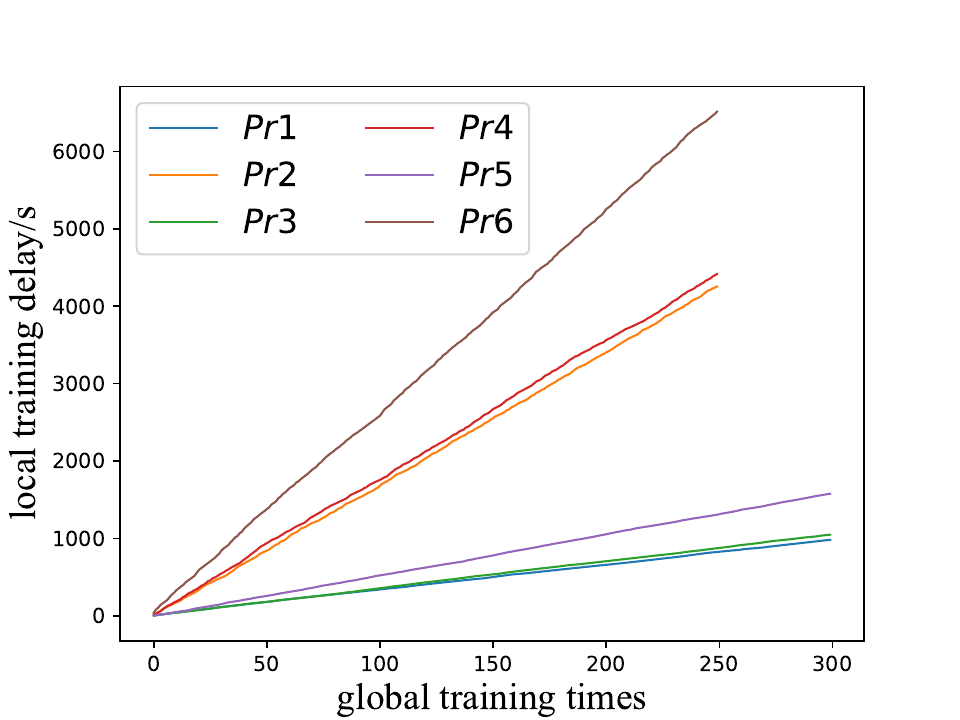} \label{fig5(c)} 
}
\caption{Results of communication performance with communication efficiency optimization for CNC  under different settings}\label{fig5}
\end{figure*}

\begin{figure*}[htbp]
\centering
\subfigure[transmission energy ]{
\includegraphics[scale=0.3]{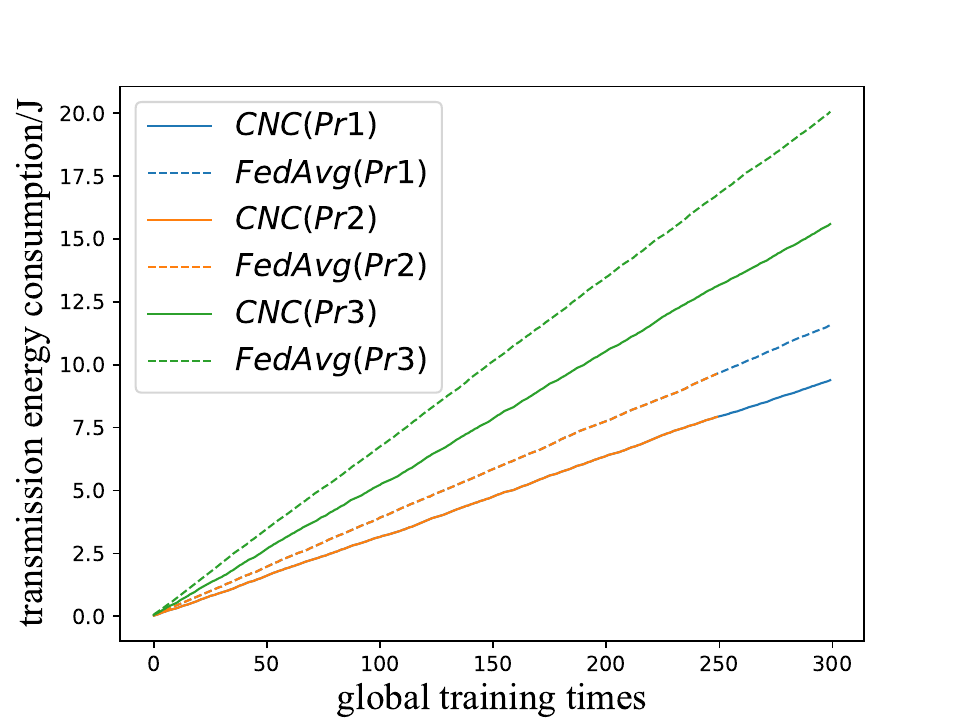} \label{fig6(a)}
}
\quad
\subfigure[transmission delay]{
\includegraphics[scale=0.3]{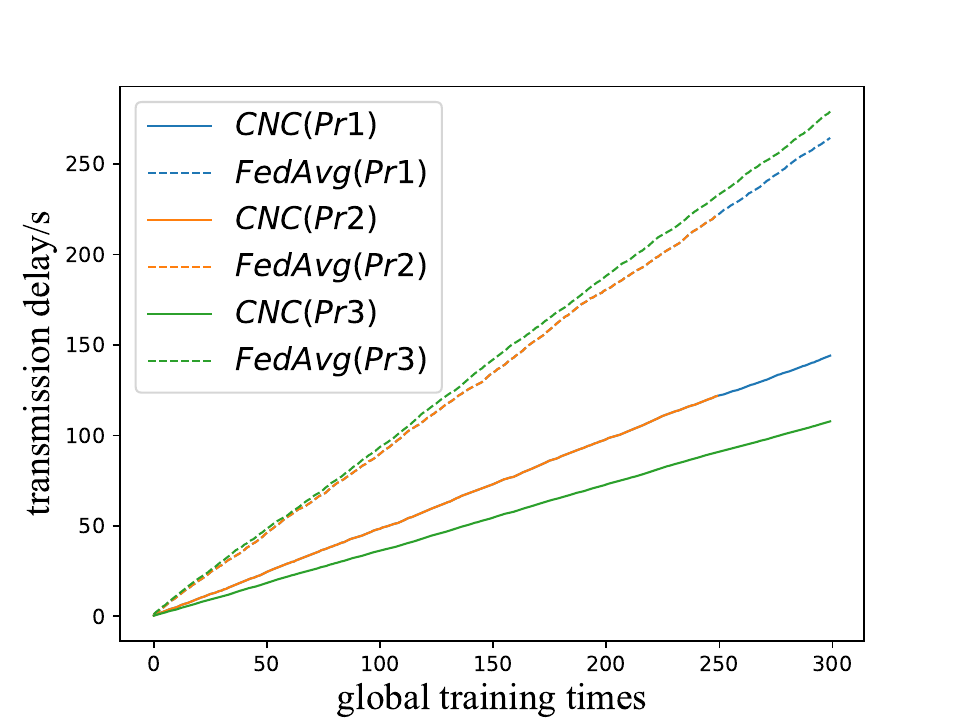} \label{fig6(b)} 
}
\quad
\subfigure[local training delay]{
\includegraphics[scale=0.3]{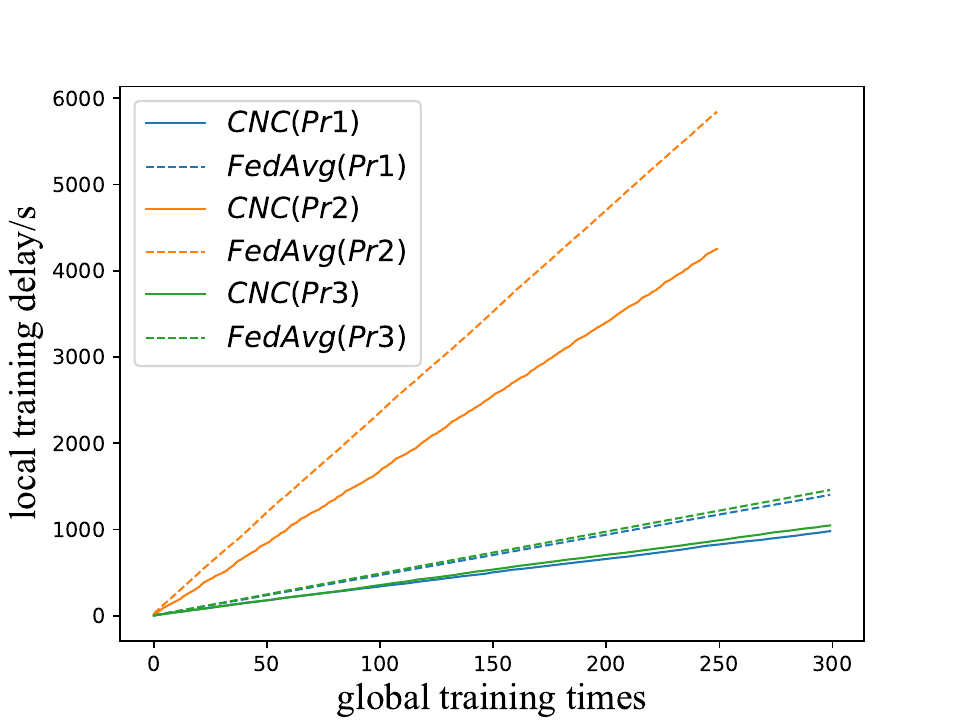} \label{fig6(c)} 
}
\caption{Results of communication performance comparison under the two algorithms}\label{fig6}
\end{figure*}

\begin{figure*}[htbp]
\centering
\subfigure[transmission energy(IID) ]{
\centering\includegraphics[scale=0.3]{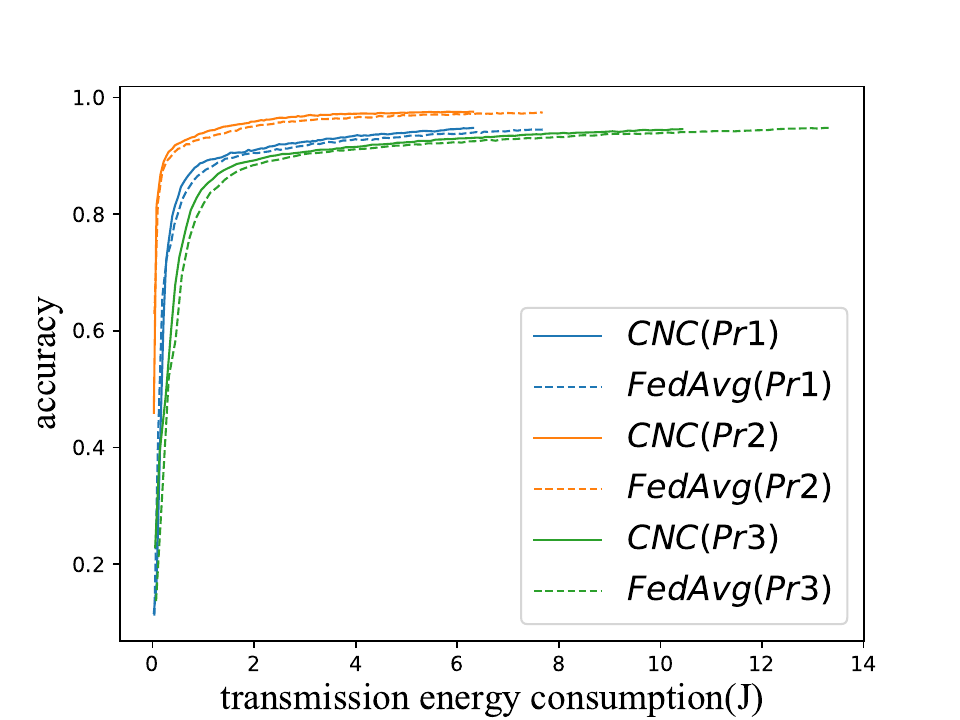} \label{fig7(a)}
}
\quad
\subfigure[transmission delay(IID)]{
\centering\includegraphics[scale=0.3]{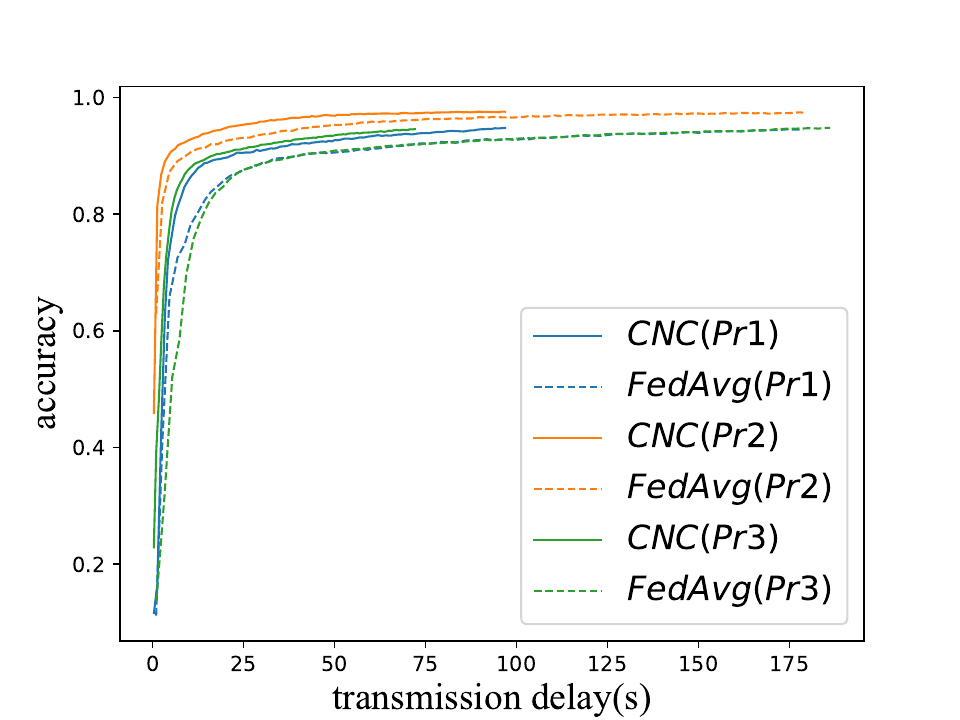} \label{fig7(b)} 
}
\quad
\subfigure[local training delay(IID)]{
\centering\includegraphics[scale=0.3]{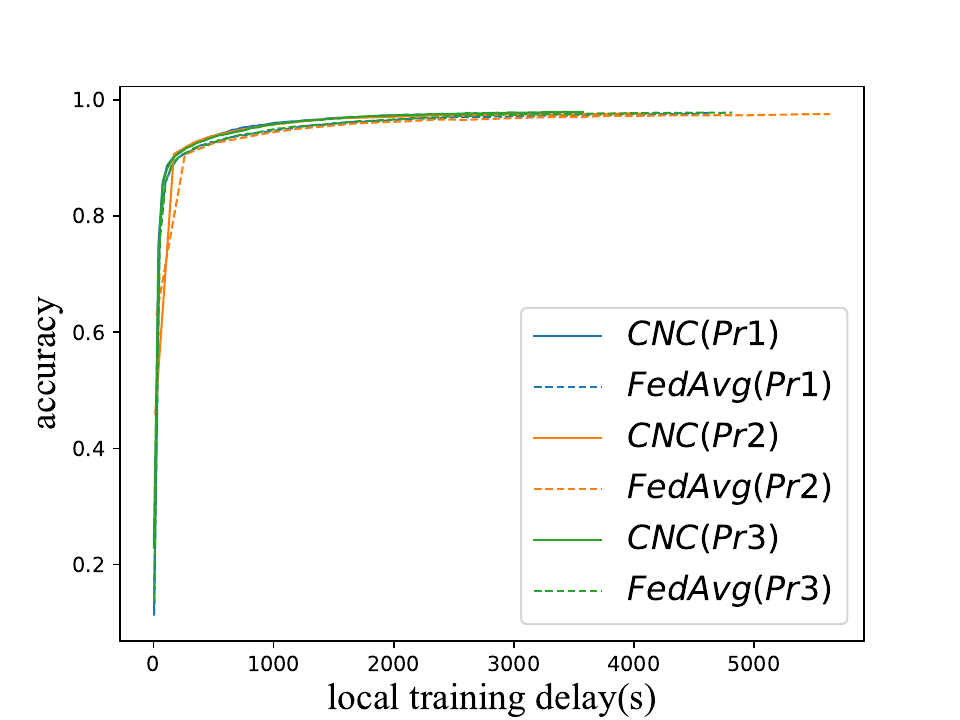} \label{fig7(c)} 
}
\quad
\subfigure[transmission energy(Non-IID)]{
\centering\includegraphics[scale=0.3]{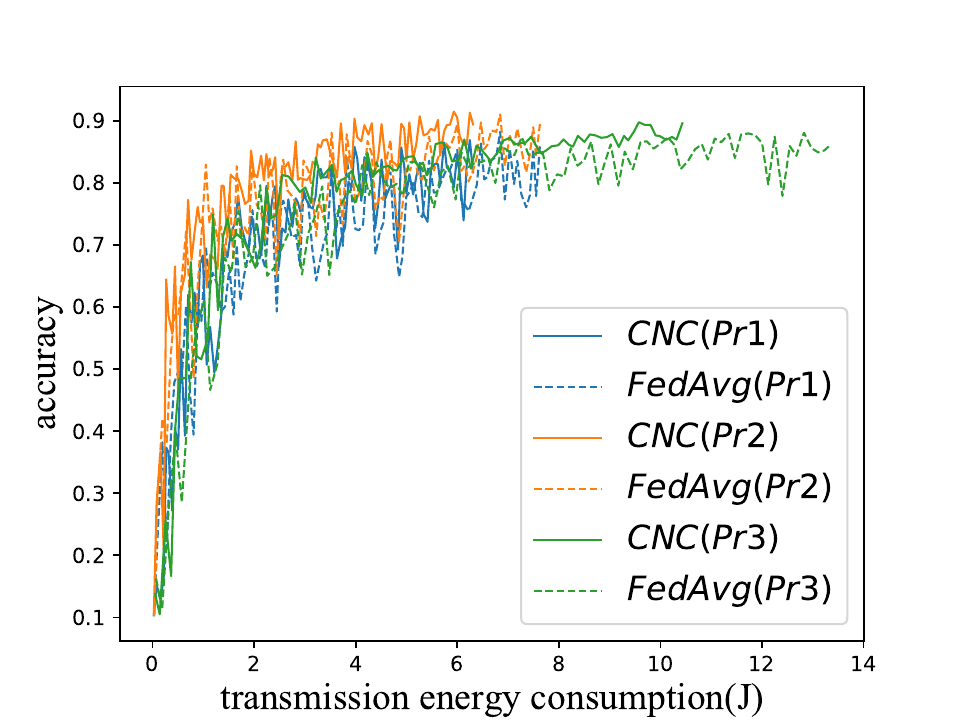} \label{fig7(d)} 
}
\quad
\subfigure[transmission delay(Non-IID)]{
\centering\includegraphics[scale=0.3]{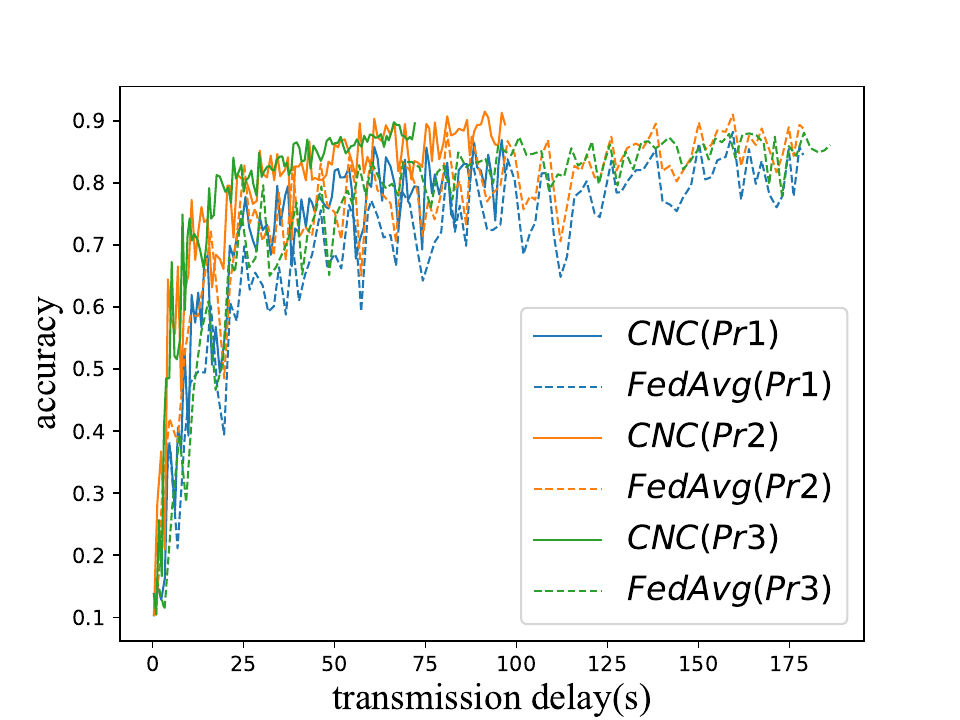} \label{fig7(e)} 
}
\quad
\subfigure[local training delay(Non-IID)]{
\centering\includegraphics[scale=0.3]{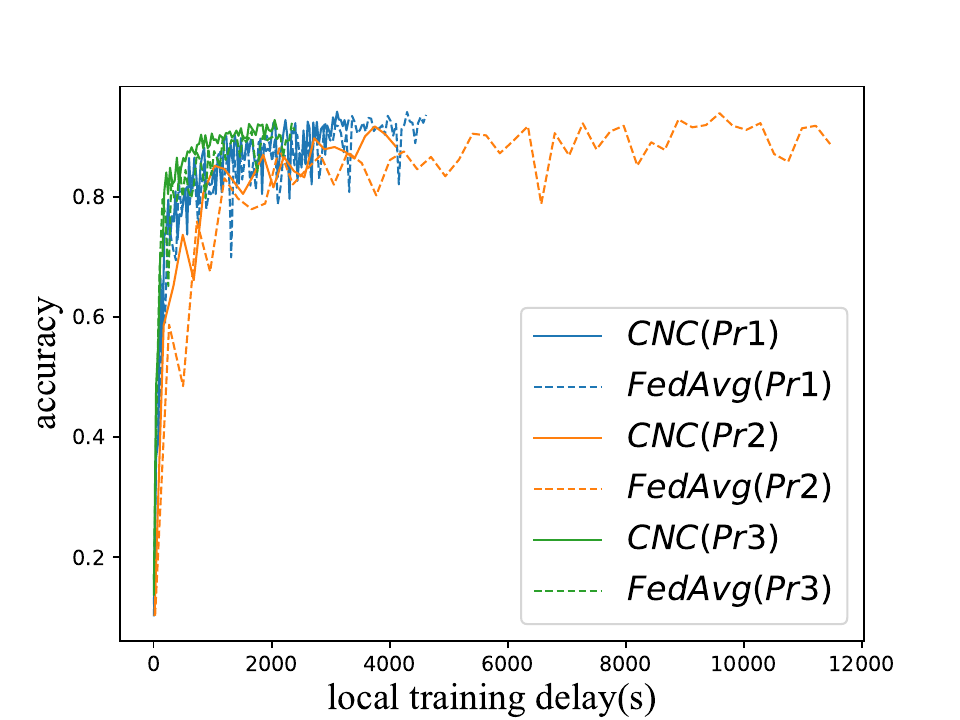} \label{fig7(f)} 
}

\caption{Results of test accuracy comparison under the two algorithms. (The horizontal axis is the different communication consumption)}\label{fig7}
\end{figure*}

Fig. \ref{fig4} shows the global model convergence curve in the CNC optimization method under the traditional architecture, with the horizontal axis indicating the number of global training. Firstly, we can see that the test accuracy converges faster and the final model accuracy is higher when the dataset is IID.

In addition, the model convergence shows different trends in different cases. Comparing the optimization of CNC in $Pr1$ within $Pr2$, the accuracy improvement is to be accelerated when the number of local training is increased. But this is at the expense of more local training time. Similarly, comparing $Pr1$ with $Pr3$, when the sampling ratio increases, the global model accuracy in $Pr3$ rises faster than in $Pr1$ as the number of global training increases. This is because under the setting of $Pr3$, there are more clients involved in training. The global model gains more gradient information each global training. 
Under settings of $Pr5$, although there are fewer sampled clients than in $Pr1$,  there is a higher amount of local data in each client. Richer information about the gradient of each global training results in a faster convergence rate than in $Pr1$.

Fig. \ref{fig5} shows the variation of the federated learning communication performance metrics under our proposed algorithm. The horizontal axis indicates the number of global training. Similar to the above description, accelerating the rate at which accuracy increases with the number of global training is achieved at the expense of some communication performance. Increasing the sampling ratio will exacerbate the local training energy consumption and the energy consumption for transmitting model parameters. Moreover, increasing the number of local training times per global training will result in longer local training delays. With the optimization of federated learning for the CNC, we can adjust the parameters set to achieve the results we want, depending on the objectives of the federated learning training.

Compared with other algorithms, the performance of our proposed algorithm is significant. As shown in Fig. \ref{fig6}, the variation curves of each performance metric with the increasing number of global training reflect the strengths of our approach. In three settings of $Pr1$,$Pr2$ and $Pr3$, optimization of CNC requires lower local training latency, lower latency to transmit the model, and lower energy consumption compared to FedAvg \cite{b5} for the whole training.
For a further comparison, we plot the variation curves of global model accuracy whereas the horizontal axis is the different communication consumption. Communication consumption include
local training delay, transmission energy consumption, and transmission delay. 
For the three cases based on the Non-IID dataset or IID dataset, as shown in Fig. \ref{fig7}. The global model accuracy of the optimization in federated learning for the CNC rises faster at the expense of the same communication consumption than FedAvg. 

In Fig. \ref{fig7(a)} and Fig. \ref{fig7(d)}, firstly the performance of our method is better than FedAvg at all three settings. Moreover, when using the Non-IID dataset, increasing the number of samples per global training can improve the convergence rate of the model as the transmission energy consumption grows. It can be seen by the change of the blue and green curves in the two plots. In Fig. \ref{fig7(b)}, when the indicator becomes transmission delay, with the IID dataset and settings of $Pr3$, our optimization approach maintains the best performance. In Fig. \ref{fig7(e)}, however, when using the Non-IID dataset, our method in settings of $Pr2$ has the best performance compared to the other five curves. This means that our optimization method of CNC is able to reduce transmission delay or energy consumption for a larger number of sampled clients. At last, in Fig. \ref{fig7(c)}, all curves show similar performance. But it can still be seen that our approach has a slight advantage over FedAvg. In Fig. \ref{fig7(f)}, while using the Non-IID dataset, the contrast between the two methods is more obvious. After the same local training delay, our method can get a model with higher accuracy. In addition, increasing the number of local training times does not necessarily speed up the convergence rate of the model, as shown by the orange line.

No matter which communication performance metric is compared with FedAvg, our method performs better. This indicates that the method we proposed can indeed optimize communication efficiency in federated learning.

\begin{figure}[htbp]
    \includegraphics[width=\linewidth,scale=0.2]{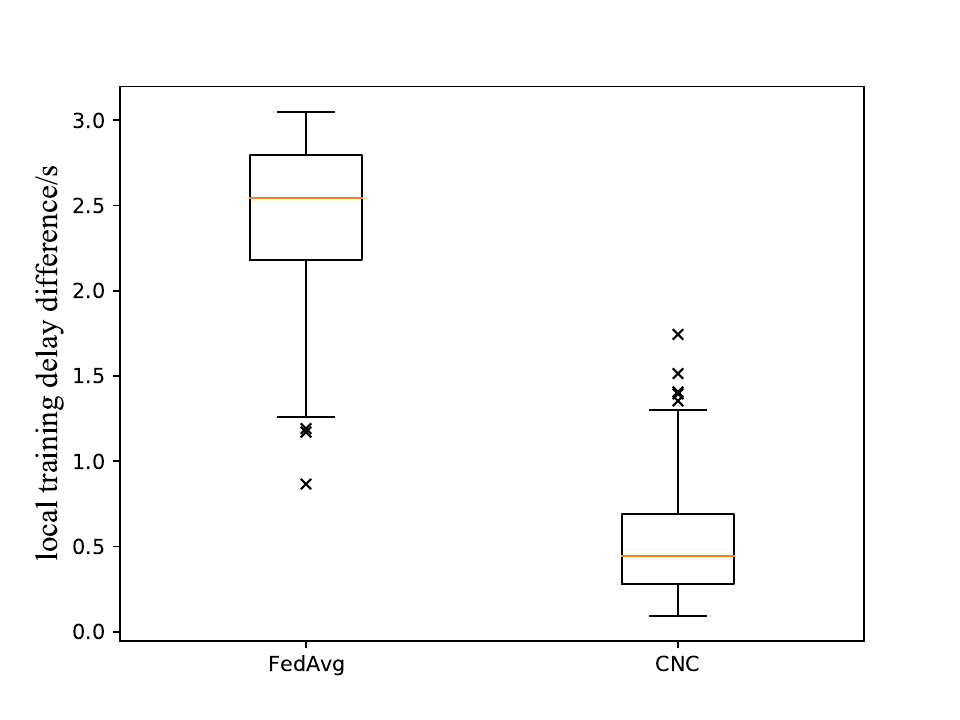}
    \caption{local training time differences in the two algorithms under the setting of $Pr1$}
    \label{fig8}
\end{figure}

At last, the box plot of the local training time differences in the two algorithms is shown in Fig. \ref{fig8}. Compared to FedAvg, our proposed method guarantees a more stable and smaller time difference.

\subsection{Simulation experiments under peer-to-peer architecture}
\subsubsection{Simulation environment parameter setting}

Due to the chain transmission of model parameters in peer-to-peer architecture, the communication time cost is huge. Time delay includes local training time and transmission time. Based on this, we designed two experiments. The first experiment has 20 clients while the second has only 8 clients.
Experiment 1 mainly compares the effect of scheduling based on computing power under optimization of CNC while experiment 2 mainly compares the performance of different path selection strategies under a few points.

In the first experiment, we designed the transmission consumption matrix of 20 clients. The numerical value represents the relative size. And the local training delay follows the assumption under the traditional architecture. All of the selection of transmission path is based on the algorithm \ref{algoritem3}. Four settings are simulated in the experiment, as follows:
\begin{enumerate}
\item According to algorithm \ref{algoritem2}, in each global training, 20 clients are divided into 4 parts on average;
\item In each global training, 20 clients are divided into 2 parts on average;
\item Each global training randomly selects 15 clients  and outputs a global model;
\item Each global training selects 20 clients and outputs a global model.
\end{enumerate}

In the second experiment, the number of clients is reduced. In order to ensure the convergence rate of the model, the optimization method of CNC chooses to divide the whole into two parts, in which the computing power resources of the main part are superior to the other part. We designed the transmission consumption matrix of 8 clients. And three different parameter settings are simulated, among which the selection of transmission path is slightly different:
\begin{enumerate}
\item All of the 8 clients participate in the training. And the transmission problem is transformed into Total Suspended Particulate (TSP) problem;
\item According to the  algorithm \ref{algoritem2}, the 8 clients are divided into two parts, the main part includes 6 clients;
\item Each global training randomly selects 6 clients.
\end{enumerate}

\subsubsection{experiment results and analysis}
\begin{figure*}[htbp]
\centering
\subfigure[local training delay(IID) ]{
\centering
\begin{minipage}[t]{0.35\linewidth}
			\centering
	\includegraphics[width=1\linewidth]{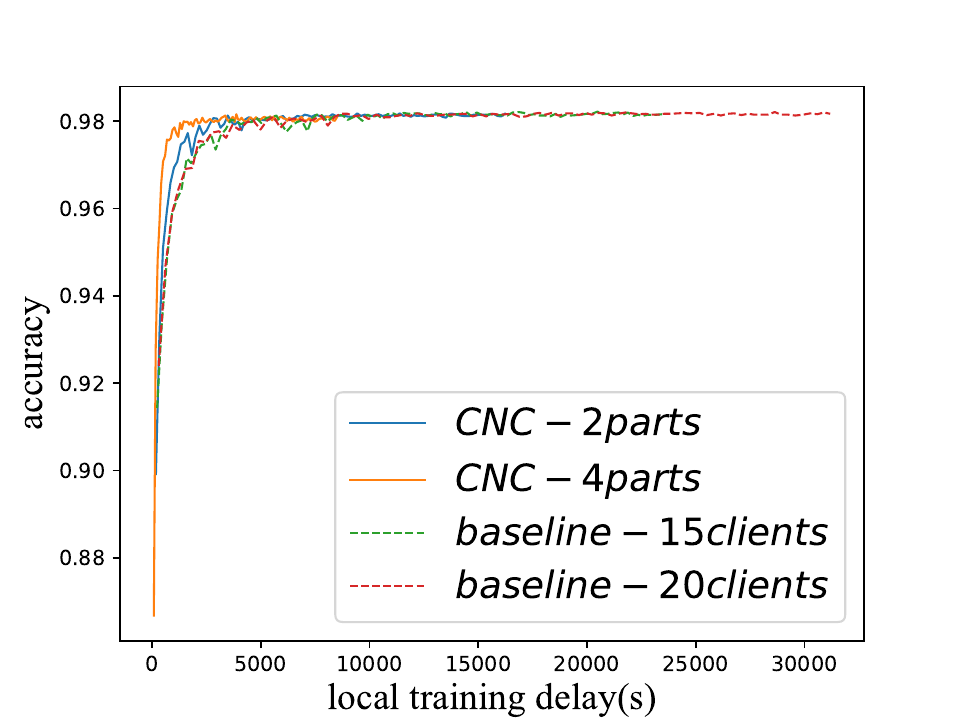}
		\end{minipage} \label{fig9(a)}
}
\quad
\subfigure[transmission consumption(IID)]{
\centering
\begin{minipage}[t]{0.35\linewidth}
			\centering
	\includegraphics[width=1\linewidth]{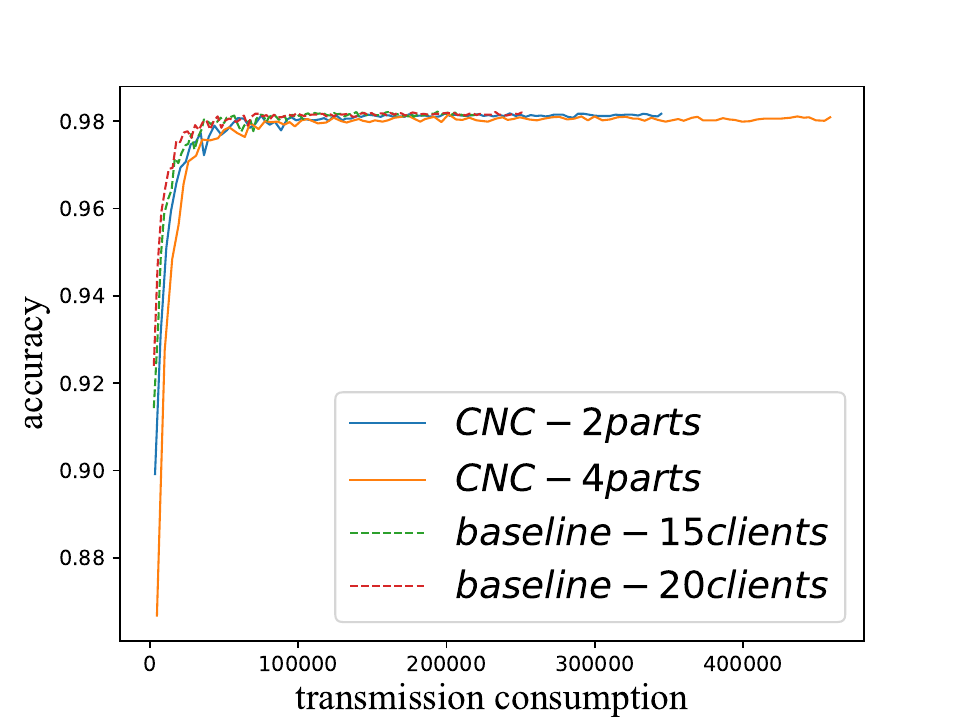}
		\end{minipage} \label{fig9(b)}
}
\quad
\subfigure[local training delay(Non-IID)]{
\centering
\begin{minipage}[t]{0.35\linewidth}
			\centering
	\includegraphics[width=1\linewidth]{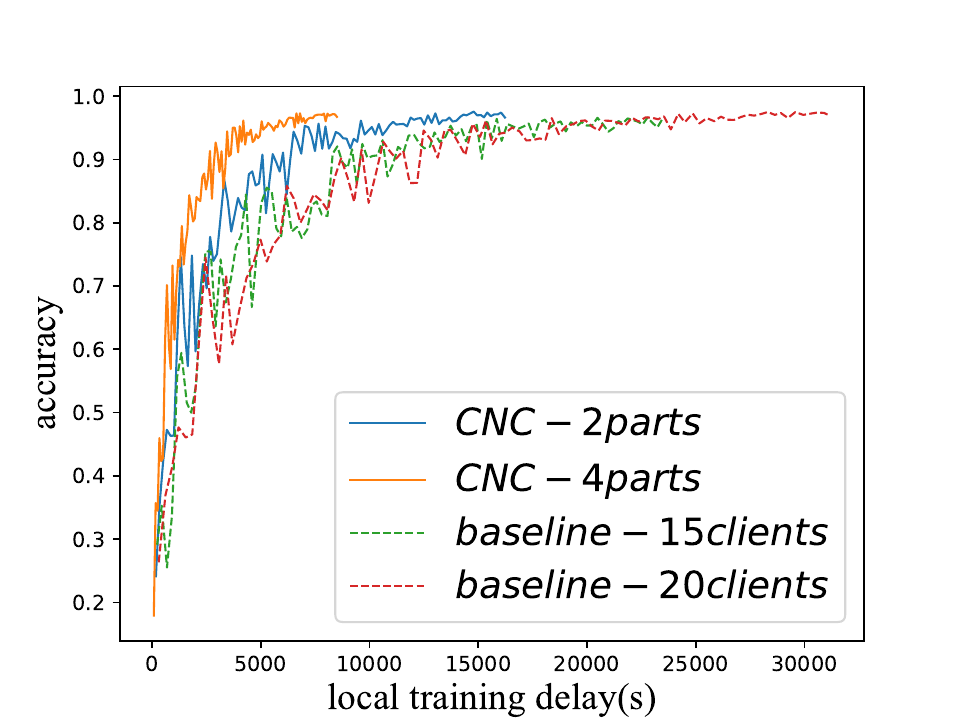}
		\end{minipage} \label{fig9(c)}
}
\quad
\subfigure[transmission consumption         (Non-IID)]{
\centering
\begin{minipage}[t]{0.35\linewidth}
			\centering
	\includegraphics[width=1\linewidth]{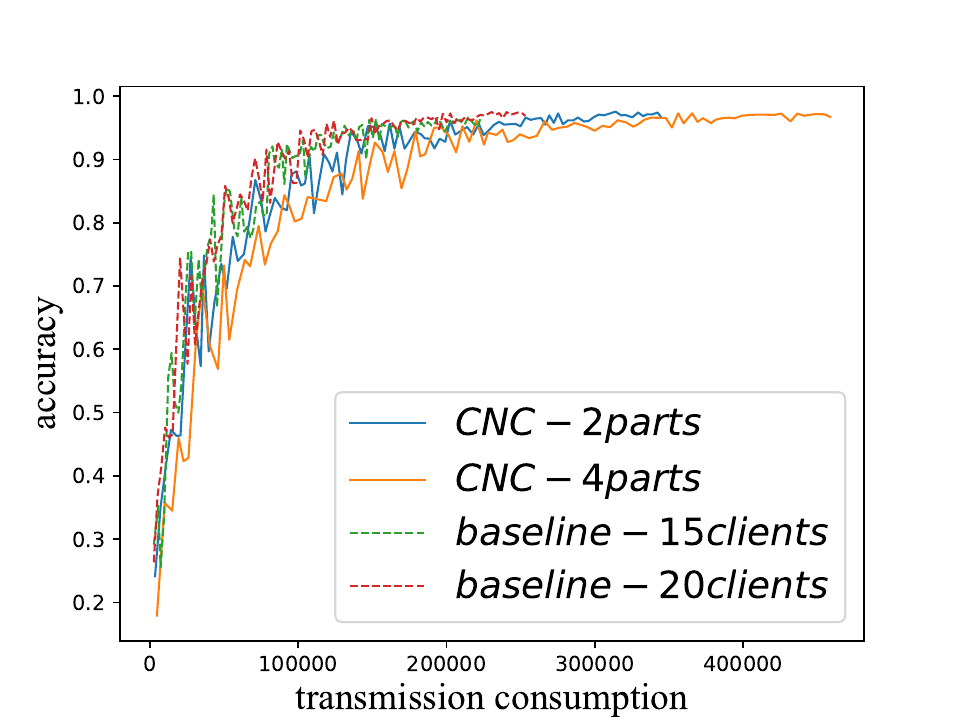}
		\end{minipage} \label{fig9(d)}
}

\caption{Results of test accuracy comparison under peer-to-peer architecture in experiment 1. (The horizontal axis is the different communication consumption)}\label{fig9}
\end{figure*}

\begin{figure*}[htbp]
\centering
\subfigure[local training delay(IID) ]{
\centering
\begin{minipage}[t]{0.35\linewidth}
			\centering
	\includegraphics[width=1\linewidth]{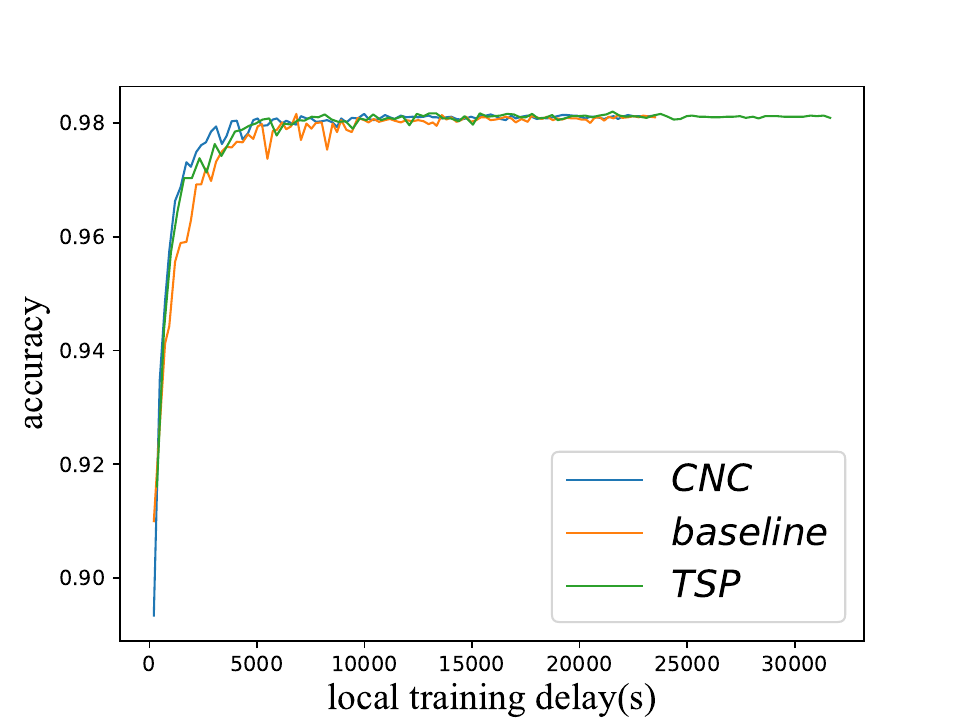}
		\end{minipage} \label{fig10(a)}
}
\quad
\subfigure[transmission consumption(IID)]{
\centering
\begin{minipage}[t]{0.35\linewidth}
			\centering
	\includegraphics[width=1\linewidth]{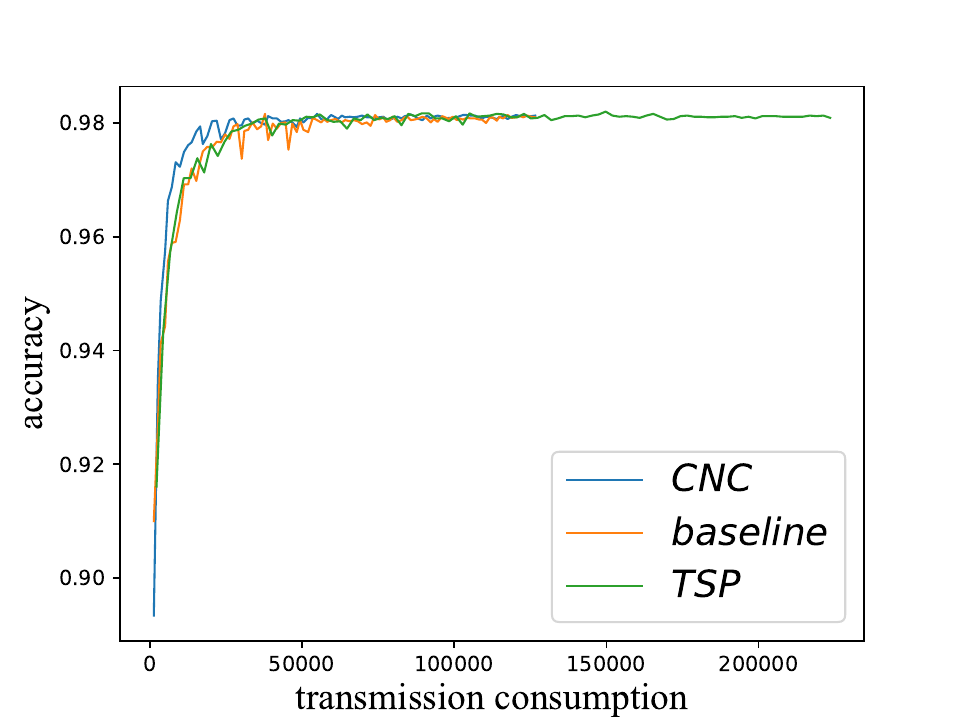}
		\end{minipage}\label{fig10(b)}
}
\quad
\subfigure[local training delay(Non-IID)]{
\begin{minipage}[t]{0.35\linewidth}
			\centering
	\includegraphics[width=1\linewidth]{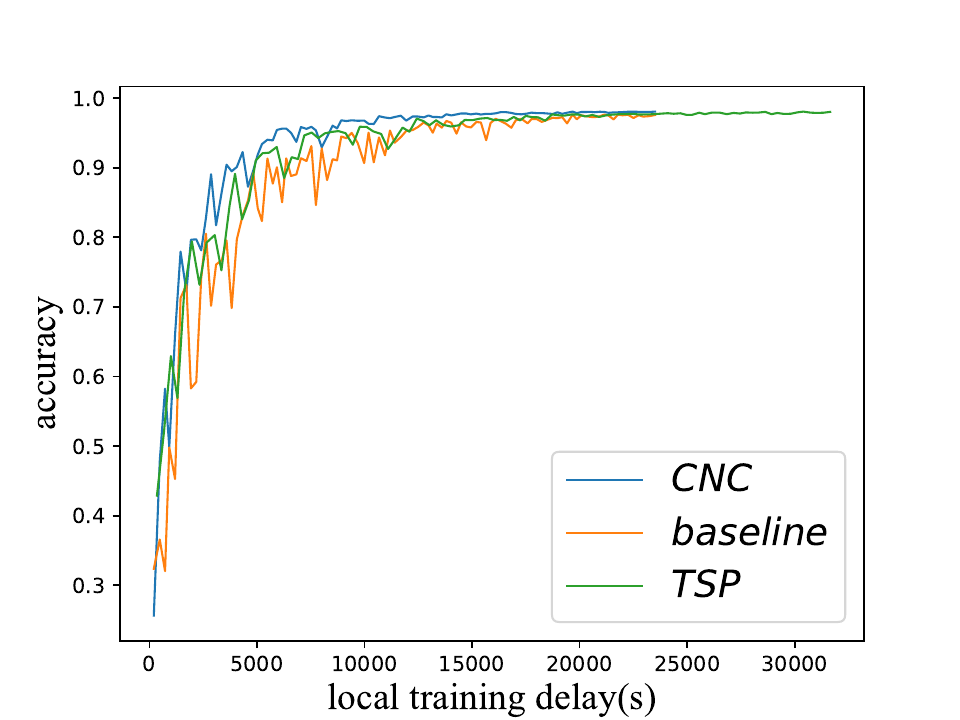}
		\end{minipage}\label{fig10(c)}
}
\quad
\subfigure[transmission consumption         
 (Non-IID)]{
\begin{minipage}[t]{0.35\linewidth}
		\centering
	\includegraphics[width=1\linewidth]{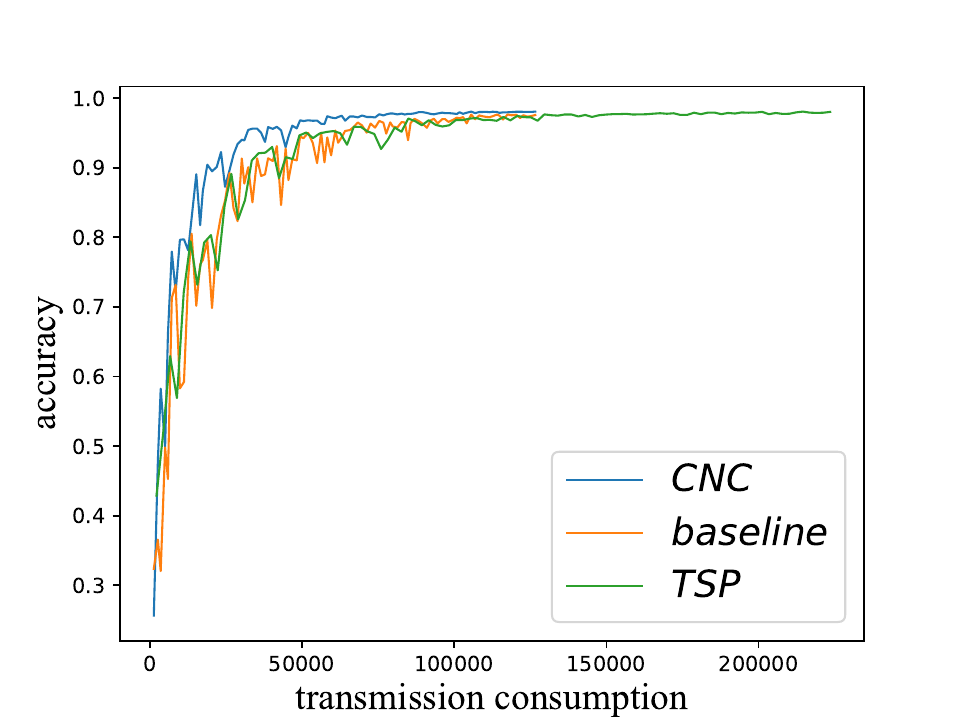}
		\end{minipage}\label{fig10(d)}
}

\caption{Results of test accuracy comparison under peer-to-peer architecture in experiment 2. (The horizontal axis is the different communication consumption)}\label{fig10}
\end{figure*}
The results of the first experiment are analyzed first.
As shown in Fig. \ref{fig9(a)} and Fig. \ref{fig9(c)}, in a peer-to-peer architecture, the dataset distribution has less impact on the algorithm. But the experimental contrast is stronger for the Non-IID dataset. When the horizontal axis is local training delay, our proposed optimization method has a faster convergence rate. And the higher the number of subsets is, the method performs better. This means our optimization method outputs a model with higher accuracy for the same local training latency. In Fig. \ref{fig9(b)} and Fig. \ref{fig9(d)}, compared to the baseline, our method needs more transmission consumption to achieve the same accuracy of the global model. But this difference is not so great as seen from the figure.  All four simulation methods use the output of Algorithm \ref{algoritem3}. The disadvantages in terms of transmission consumption are to be expected. 

Overall, Our proposed optimization method greatly reduces the local training latency during training without excessive transmission energy or latency consumption. But in reality, the cost of local training delay is often higher than transmission consumption. From this point of view, our proposed communication efficiency optimization of federated learning for CNC is more practical.

The results of the second experiment under the peer-to-peer architecture and its analysis are as follows.
In Fig. \ref{fig10(a)} and Fig. \ref{fig10(c)}, we compare algorithm performance in terms of local training time delay. The global model of optimization of CNC converges faster. Moreover, our proposed method maintains good performance when measured in terms of transmission consumption, as shown in Fig. \ref{fig10(b)} and Fig. \ref{fig10(d)}. The optimization under peer-to-peer structures can be advantageous when there are fewer nodes involved in federated learning. Solving the TSP problem does not guarantee a small local training time, although it is possible to find the optimal transmission path. And in the Baseline, although it guarantees smaller transmission consumption and local training delay, the gradient information obtained by the model per global training is not rich compared to our proposed method.
In addition, optimization of CNC in peer-to-peer architecture can output models with higher accuracy than traditional architecture. The average accuracy of the former is close to 92\% while the latter is as high as about 97\%.

\begin{figure}[htbp]
    \includegraphics[width=\linewidth,scale=0.75]{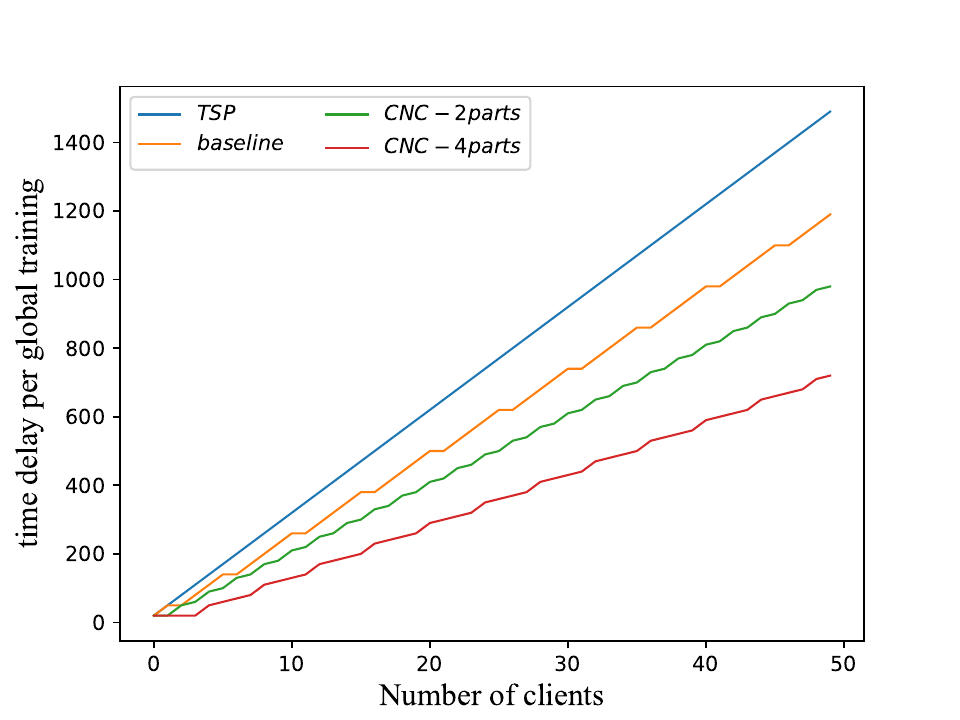}
    \caption{Variation of the average global training latency with the number of clients}
    \label{fig11}
\end{figure}

Finally, we qualitatively study the variation of the average global training latency with the number of clients
in the peer-to-peer architecture. Fig. \ref{fig11} shows the result. Compared to other methods, optimization of federated learning for CNC of 6G Networks can guarantee a lower latency rise rate. This ensures very high communication efficiency during federated training.

\section{Conclusion}

CNC of 6G Networks is a new network architecture and paradigm. Federated Learning can show better performance with its support, especially in terms of communication efficiency. In this case, we proposed communication efficiency optimization of federated learning for Computing and Network Convergence of 6G Networks. As the results of the experiments show, it has great potential to improve the federated learning process, especially in balancing heterogeneous computing power and improving communication efficiency.

\EOD


\begin{thebibliography}{10}
\providecommand{\url}[1]{#1}
\csname url@samestyle\endcsname
\providecommand{\newblock}{\relax}
\providecommand{\bibinfo}[2]{#2}
\providecommand{\BIBentrySTDinterwordspacing}{\spaceskip=0pt\relax}
\providecommand{\BIBentryALTinterwordstretchfactor}{4}
\providecommand{\BIBentryALTinterwordspacing}{\spaceskip=\fontdimen2\font plus
\BIBentryALTinterwordstretchfactor\fontdimen3\font minus
  \fontdimen4\font\relax}
\providecommand{\BIBforeignlanguage}[2]{{%
\expandafter\ifx\csname l@#1\endcsname\relax
\typeout{** WARNING: IEEEtran.bst: No hyphenation pattern has been}%
\typeout{** loaded for the language `#1'. Using the pattern for}%
\typeout{** the default language instead.}%
\else
\language=\csname l@#1\endcsname
\fi
#2}}
\providecommand{\BIBdecl}{\relax}
\BIBdecl

\bibitem{b1}
P.~Kairouz, H.~B. McMahan, B.~Avent, A.~Bellet, M.~Bennis, A.~N. Bhagoji,
  K.~Bonawitz, Z.~Charles, G.~Cormode, R.~Cummings \emph{et~al.}, ``Advances
  and open problems in federated learning,'' \emph{Foundations and
  Trends{\textregistered} in Machine Learning}, vol.~14, no. 1--2, pp. 1--210,
  2021.

\bibitem{b2}
O.~A. Wahab, A.~Mourad, H.~Otrok, and T.~Taleb, ``Federated machine learning:
  Survey, multi-level classification, desirable criteria and future directions
  in communication and networking systems,'' \emph{IEEE Communications Surveys
  \& Tutorials}, vol.~23, no.~2, pp. 1342--1397, 2021.

\bibitem{b3}
T.~Li, A.~K. Sahu, A.~Talwalkar, and V.~Smith, ``Federated learning:
  Challenges, methods, and future directions,'' \emph{IEEE Signal Processing
  Magazine}, vol.~37, no.~3, pp. 50--60, 2020.

\bibitem{b4}
J.~Kone{\v{c}}n{\`y}, H.~B. McMahan, F.~X. Yu, P.~Richt{\'a}rik, A.~T. Suresh,
  and D.~Bacon, ``Federated learning: Strategies for improving communication
  efficiency,'' \emph{arXiv preprint arXiv:1610.05492}, 2016.

\bibitem{b5}
B.~McMahan, E.~Moore, D.~Ramage, S.~Hampson, and B.~A. y~Arcas,
  ``Communication-efficient learning of deep networks from decentralized
  data,'' in \emph{Artificial intelligence and statistics}.\hskip 1em plus
  0.5em minus 0.4em\relax PMLR, 2017, pp. 1273--1282.

\bibitem{b6}
Y.~Fraboni, R.~Vidal, L.~Kameni, and M.~Lorenzi, ``Clustered sampling:
  Low-variance and improved representativity for clients selection in federated
  learning,'' in \emph{International Conference on Machine Learning}.\hskip 1em
  plus 0.5em minus 0.4em\relax PMLR, 2021, pp. 3407--3416.

\bibitem{b7}
W.~Wu, L.~He, W.~Lin, R.~Mao, C.~Maple, and S.~Jarvis, ``Safa: A
  semi-asynchronous protocol for fast federated learning with low overhead,''
  \emph{IEEE Transactions on Computers}, vol.~70, no.~5, pp. 655--668, 2020.

\bibitem{b8}
H.~H. Yang, Z.~Liu, T.~Q. Quek, and H.~V. Poor, ``Scheduling policies for
  federated learning in wireless networks,'' \emph{IEEE transactions on
  communications}, vol.~68, no.~1, pp. 317--333, 2019.

\bibitem{b9}
J.~So, B.~G{\"u}ler, and A.~S. Avestimehr, ``Turbo-aggregate: Breaking the
  quadratic aggregation barrier in secure federated learning,'' \emph{IEEE
  Journal on Selected Areas in Information Theory}, vol.~2, no.~1, pp.
  479--489, 2021.

\bibitem{b10}
L.~Liu, J.~Zhang, S.~Song, and K.~B. Letaief, ``Client-edge-cloud hierarchical
  federated learning,'' in \emph{ICC 2020-2020 IEEE International Conference on
  Communications (ICC)}.\hskip 1em plus 0.5em minus 0.4em\relax IEEE, 2020, pp.
  1--6.

\bibitem{b11}
Y.~Deng, F.~Lyu, J.~Ren, Y.~Zhang, Y.~Zhou, Y.~Zhang, and Y.~Yang, ``Share:
  Shaping data distribution at edge for communication-efficient hierarchical
  federated learning,'' in \emph{2021 IEEE 41st International Conference on
  Distributed Computing Systems (ICDCS)}.\hskip 1em plus 0.5em minus
  0.4em\relax IEEE, 2021, pp. 24--34.

\bibitem{b12}
T.~Lin, L.~Kong, S.~U. Stich, and M.~Jaggi, ``Ensemble distillation for robust
  model fusion in federated learning,'' \emph{Advances in Neural Information
  Processing Systems}, vol.~33, pp. 2351--2363, 2020.

\bibitem{b13}
C.~He, M.~Annavaram, and S.~Avestimehr, ``Group knowledge transfer: Federated
  learning of large cnns at the edge,'' \emph{Advances in Neural Information
  Processing Systems}, vol.~33, pp. 14\,068--14\,080, 2020.

\bibitem{b14}
D.~Li and J.~Wang, ``Fedmd: Heterogenous federated learning via model
  distillation,'' \emph{arXiv preprint arXiv:1910.03581}, 2019.

\bibitem{b15}
C.~T. Dinh, N.~H. Tran, M.~N. Nguyen, C.~S. Hong, W.~Bao, A.~Y. Zomaya, and
  V.~Gramoli, ``Federated learning over wireless networks: Convergence analysis
  and resource allocation,'' \emph{IEEE/ACM Transactions on Networking},
  vol.~29, no.~1, pp. 398--409, 2020.

\bibitem{b16}
M.~Chen, Z.~Yang, W.~Saad, C.~Yin, H.~V. Poor, and S.~Cui, ``A joint learning
  and communications framework for federated learning over wireless networks,''
  \emph{IEEE Transactions on Wireless Communications}, vol.~20, no.~1, pp.
  269--283, 2020.

\bibitem{b17}
Z.~Qin, G.~Y. Li, and H.~Ye, ``Federated learning and wireless
  communications,'' \emph{IEEE Wireless Communications}, vol.~28, no.~5, pp.
  134--140, 2021.

\bibitem{b18}
Q.~Cao, X.~Zhang, Y.~Zhang, and Y.~Zhu, ``Layered model aggregation based
  federated learning in mobile edge networks,'' in \emph{2021 IEEE/CIC
  International Conference on Communications in China (ICCC)}.\hskip 1em plus
  0.5em minus 0.4em\relax IEEE, 2021, pp. 1--6.

\bibitem{b19}
P.~Pinyoanuntapong, P.~Janakaraj, P.~Wang, M.~Lee, and C.~Chen, ``Fedair:
  Towards multi-hop federated learning over-the-air,'' in \emph{2020 IEEE 21st
  International Workshop on Signal Processing Advances in Wireless
  Communications (SPAWC)}.\hskip 1em plus 0.5em minus 0.4em\relax IEEE, 2020,
  pp. 1--5.

\bibitem{b20}
Y.~Sun, J.~Liu, H.~Huang, X.~Zhang, B.~Lei, J.~Peng, and W.~Wang, ``Computing
  power network: A survey,'' \emph{to appear in China Communications, online in arXiv}, 2022.

\end{thebibliography}
\end{document}